\documentclass[letterpaper]{article} 
\usepackage{aaai25} 
\usepackage{times}  
\usepackage{helvet}  
\usepackage{courier}  
\usepackage[hyphens]{url}  
\usepackage{graphicx} 
\usepackage{amsfonts}
\usepackage{natbib}  
\usepackage{caption} 
\usepackage{algorithm}
\usepackage{algorithmic}
\usepackage{xcolor}
\usepackage{amsfonts}
\usepackage{multirow}
\usepackage{booktabs}
\def\dataset{\textsc{PatentDesc-355K}}
\def\model{\textsc{PatentLMM}}

\usepackage{newfloat}
\usepackage{listings}
\DeclareCaptionStyle{ruled}{labelfont=normalfont,labelsep=colon,strut=off} 
\floatstyle{ruled}
\newfloat{listing}{tb}{lst}{}
\floatname{listing}{Listing}
\title{\model{}: {L}arge {M}ultimodal {M}odel for\\ Generating Descriptions for {Patent} Figures}
\author{
Shreya Shukla\textsuperscript{\rm 1}\equalcontrib,~Nakul Sharma\textsuperscript{\rm 1}\equalcontrib,   ~Manish Gupta\textsuperscript{\rm 2},~Anand Mishra\textsuperscript{\rm 1}
}
\affiliations{
    \textsuperscript{\rm 1}Indian Institute of Technology Jodhpur, India \\
    \textsuperscript{\rm 2}Microsoft, India\\
    \{shukla.12,sharma.86\}@iitj.ac.in, gmanish@microsoft.com, mishra@iitj.ac.in
}

\usepackage{bibentry}

\begin{document}

\maketitle

\begin{abstract}
Writing comprehensive and accurate descriptions of technical drawings in patent documents is crucial to effective knowledge sharing and enabling the replication and protection of intellectual property. 
However, automation of this task has been largely overlooked by the research community. To this end, we introduce \dataset{}, a novel large-scale dataset containing $\sim$355K patent figures along with their brief and detailed textual descriptions extracted from 60K+ US patent documents. In addition, we propose \model{} -- a novel multimodal large language model specifically tailored to generate high-quality descriptions of patent figures. Our proposed \model{} comprises two key components: (i) \textsc{PatentMME}, a specialized multimodal vision encoder that captures the unique structural elements of patent figures, and (ii) \textsc{PatentLLaMA}, a domain-adapted version of LLaMA fine-tuned on a large collection of patents. Extensive experiments demonstrate that training a vision encoder specifically designed for patent figures significantly boosts the performance, generating coherent descriptions compared to fine-tuning similar-sized off-the-shelf multimodal models. \dataset{} and \model{} pave the way for automating the understanding of patent figures, enabling efficient knowledge sharing and faster drafting of patent documents. We make the code and data publicly available\footnote{https://vl2g.github.io/projects/PatentLMM/}.
\end{abstract}

%

\section{Introduction}
\label{sec:intro}
Patents are a cornerstone of intellectual property protection, granting inventors exclusive rights to their creations. Effective communication of these inventions is crucial for patent examiners, courts, and the technical community to appreciate the inventiveness of these inventions and assess their novelty. Patent documents rely heavily on figures and their corresponding textual descriptions to present technical details. Writing accurate descriptions of these figures is essential for an unambiguous understanding of the invention and its components and facilitates knowledge sharing within the technical community. Comprehensive descriptions also ensure that the invention is adequately protected against potential infringements by others. However, manually crafting such descriptions is time-consuming and laborious, hindering the efficiency of patent processing and analysis.

One of the major challenges for generating patent figure descriptions in an automated way is the lack of large-scale labeled datasets. 
Existing datasets, while invaluable for advancing research in natural and scientific figure captioning, do not adequately capture the nuances and complexities inherent to patent illustrations. To address this gap, we curate \dataset{}, a novel large-scale dataset containing $\sim$355K patent figures and their brief and detailed textual descriptions extracted from 60K+ patent documents. This dataset offers a rich and diverse collection of patent figures that span various technical domains, along with their corresponding descriptions, enabling the development and evaluation of models specifically tailored for this task.

Typically, patent figures are associated with brief and detailed descriptions. In our proposed \dataset{} dataset, we found that they span an average of $\sim$34 and $\sim$1680 tokens, respectively. Thus, unlike existing image captioning benchmarks, for example COCO~\cite{Lin2014mscoco}, TextCaps~\cite{sidorov2020textcaps} and NoCaps~\cite{agrawal2019nocaps} where captions span an average of $\sim$12 tokens, the descriptive captioning of patent figures in our dataset is much more challenging. Moreover, unlike the natural scene images of the existing captioning datasets, patent figures are structured technical illustrations that adhere to a more standardized visual style for technical and legal documentation.

\begin{figure*}[!t]
\centering
\includegraphics[width=\textwidth]{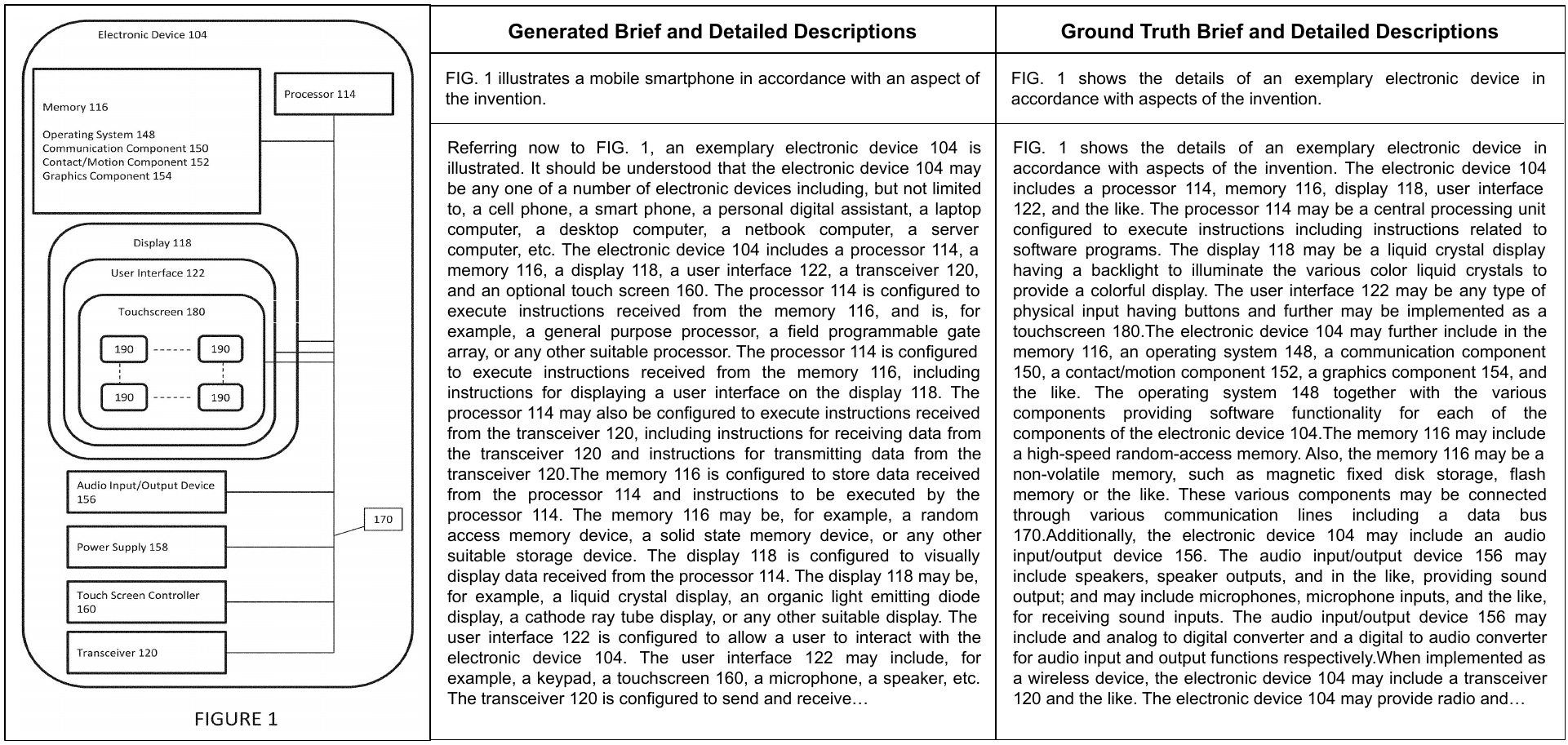}
\caption{\label{fig:qual_res}An example of generated and ground truth brief and detailed descriptions using our proposed \model{}.}

\end{figure*} 

The emergence of Large Language Models (LLMs) and Large Multimodal Models (LMMs) has revolutionized almost every vision and language task. These models exhibit a remarkable ability to understand and generate coherent language across diverse domains. However, applying these models to the generation of patent descriptions presents unique challenges. The length of descriptions and the complexity inherent to patent diagrams underscore the need to focus on various elements of the figure, such as arrows, nodes, and text annotations. Further, contrary to dense document images, patent figures are sparse and comprise several elements like text, nodes, node labels (a number associated with nodes in the patent figure), figure numbers, and arrows in different styles, i.e., uni-direction and bidirectional, solid, and dotted, among others. Please refer to Fig. 3 in the Appendix for an overview of these elements.

As can be seen in Fig.~\ref{fig:qual_res}, the detailed description of patent figures heavily makes use of these elements to convey the semantics of the figure. Given this dramatic difference between captions of natural scenes versus patent figures, it was anticipated that recent image captioning methods~\cite{Li2022BLIPBL,Wang2022GITAG,OFA} and multimodal LLMs~\cite{ye2023mplug-owl, LLaVA-1.5, Zhu2023MiniGPT4EV} would perform poorly for our task in a zero-shot setting. Surprisingly, these approaches demonstrated suboptimal performance even after fine-tuning on our dataset. These unique properties of patent figures require specialized system design to ensure the accurate and concise generation of descriptions without introducing hallucinations or irrelevant details.

In this paper, we propose \model{} -- a novel model to generate descriptions of patent figures. The model contains two important components: \textsc{PatentMME} and \textsc{PatentLLaMA}. \textsc{PatentMME} is a specialized multi-modal vision encoder for patent figures, trained using masked language modeling loss, along with two other novel loss functions focused on learning structure from sparse patent figures. \textsc{PatentLLaMA} is a domain-adapted version of LLaMA fine-tuned on a large collection of patent text from the Harvard USPTO Dataset (HUPD)~\cite{suzgun2024harvard}. \model{} combines the \textsc{PatentMME} encoder and the \textsc{PatentLLaMA} using a projection layer. 

\noindent The major contributions of our work are as follows.
    (i) We present a large-scale dataset of $\sim$355K patent figures and their brief and detailed descriptions. 
    (ii) We propose a novel multimodal model \model{}, comprising a patent domain-specialized vision encoder trained using objectives specifically tailored to capture the structure of patent documents and an LLM fine-tuned on patent data.
    (iii) We extensively benchmark existing captioning models and multimodal LLMs and show that our proposed approach surpasses their best performance on the average BLEU metric by 10.22\% and 4.43\% on an absolute scale for generating brief and detailed descriptions, respectively.
\section{Related Work}
\label{sec:related_work}
\noindent\textbf{Image Captioning in Pre-LMMs era:}
The patent figure description task is broadly similar to the image captioning task, which has been an active research area in the last decade. Some representative early work on image captioning includes the combination of a CNN encoder with an LSTM decoder~\citep{vinyals2015show}, a multimodal RNN architecture that uses local and global image features~\citep{andreas2016neural}, an adaptive attention model~\citep{lu2017knowing}, and a bottom-up and top-down attention model~\citep{anderson2018bottom}. 
Recent works have also focused on improving caption diversity~\cite{shetty2017speaking}, novel object captioning~\cite{lu2018neural}, and incorporating external knowledge~\cite{gu2019unified}. 
As discussed in the previous section, our task differs significantly from these previous efforts on image captioning in terms of the length of descriptions and the structure of patent figures.

\noindent\textbf{Describing Scientific Figures:}
Patent figures are a specific form of scientific illustrations. Although previous work on generating descriptions of patent figures has been sparse, ample research has been done to caption scientific figures.
~\citet{neuralcaption2019,chen2020figcapreasoning} create and leverage FigCAP and adapt an LSTM-based model~\cite{lstm} for captioning. Recently,~\citet{scicap} collected the SciCap dataset from articles published on arXiv
In~\cite{scicap+}, the authors augment the SciCap dataset with additional information such as OCR text from figures and referring sentences from the text to curate SciCap+, and demonstrate the performance boost achieved by incorporating extra information.~\citet{chart-to-text} and \citet{tang2023vistext} address the problem of captioning various visualization charts of data. Certain works go beyond natural language descriptions to generate code, particularly for flowcharts. For example, ~\citet{shukla2023floco} and ~\citet{LiuHZLZX22} specifically address the generation of code from flow chart images.
A parallel work PatFig~\cite{aubakirova2023patfig} scrapes a similar dataset as ours with 17K training samples and 2K test samples, and demonstrates the performance of MiniGPT-4~\cite{Zhu2023MiniGPT4EV} in the proposed dataset. In this work, we contribute a $\sim$20$\times$ larger dataset and propose a novel model, \model{}, which is almost twice as effective as MiniGPT-4 in BLEU-4 for \dataset{}.

\noindent\textbf{Large Multimodal Models:}
Recent work in the multimodal (vision and language) community has focused on leveraging the world knowledge implicitly encoded in large language models for multimodal tasks such as visual question answering and image captioning~\cite{Zhu2023MiniGPT4EV, Li2023BLIP2BL, LLaVA-1.5, achiam2023gpt4, ye2023mplug-owl, ye2023mplug-owl2, OFA, team2023gemini, Alayrac2022FlamingoAV}, visual grounding~\cite{ye2023mplug-owl2, Zhu2023MiniGPT4EV, team2023gemini, achiam2023gpt4} and image-text matching~\cite{Li2022BLIPBL, Li2023BLIP2BL}. This is achieved by feeding an image representation as input along with the prompt to the language model and modeling the output using the language modeling objective. 
Recent advances include Flamingo~\cite{Alayrac2022FlamingoAV}, which inserts trainable gated cross-attention layers into a pretrained LLM~\cite{Hoffmann2022TrainingCLChinchilla}. 
BLIP-2~\cite{Li2023BLIP2BL} leverages pre-trained ViT~\cite{ViT} and LLaMA~\cite{LLaMA-meta}, combined with QFormer, to translate image embeddings into LLM prompt embeddings. MiniGPT-4~\cite{Zhu2023MiniGPT4EV} builds upon pretrained BLIP-2 and finetunes an additional linear layer to project queries into the LLM on a curated dataset. In contrast, LLaVA-1.5~\cite{LLaVA-1.5} proposes a relatively simple and effective two-stage approach. 
In addition, document-specific LLMs such as LayoutLLM~\cite{luo2024layoutllm}, UReader~\cite{ye2023ureader} and TextMonkey~\cite{liu2024textmonkey} 
have shown impressive performance on Document VQA task. We compare with several of these models and show that these models do not perform competitively for the task of generating descriptions from patent figures.

\section{\dataset{}: A Novel Dataset of Patent Figures with Descriptions}
\label{sec:dataset}
We introduce \dataset{} -- a novel large-scale dataset tailored for generating descriptions for patent figures. Our proposed dataset comprises 355K patent figures sourced from Google Patents\footnote{https://patents.google.com}, with each image accompanied by its brief and detailed descriptions extracted from the corresponding patent documents. The dataset is available for download on our project website:~\url{https://vl2g.github.io/projects/PatentLMM/}. Fig.~\ref{fig:qual_res} visualizes a $\langle$patent figure, brief description, detailed description$\rangle$ triplet from our dataset. With our primary focus on US patents published after 2004, our dataset spans over 60K patents from assignees like Amazon, Microsoft, LinkedIn, Google, Yahoo, etc. To assess the quality of the dataset, we manually evaluated a random set of 100 patent figures with their brief and detailed descriptions and computed the sentence-level precision and recall of the extracted descriptions against the ground-truth descriptions. For brief descriptions, both precision and recall scores were
100\%. For detailed descriptions, precision and recall were 90. 81\% and 91. 96\%, respectively. More details on data set curation, preprocessing, description extraction, and quality assessment are provided in Appendix A. 
\begin{table}[!t]
\centering
\scriptsize
\begin{tabular}{l r r r}
\toprule
&\textbf{Train} & \textbf{Validation} & \textbf{Test}\\
\midrule
Number of Images & 320,717 & 17,286 & 17,336\\
Avg. number of tokens in brief descriptions & 34.37 & 34.28 & 34.30\\
Avg. number of tokens in detailed descriptions & 1,677.85 & 1,676.71 & 1,697.16\\
Number of Unique Patents & 50,448 & 8,027 & 7,964\\
Avg. number of images per patent & 6.36 & 2.15 & 2.18\\
\bottomrule
\end{tabular}
\caption{\dataset{}: Dataset Statistics.}
\label{tab:dataset_stats}
\end{table}

\noindent\textbf{Dataset Analysis}: Table~\ref{tab:dataset_stats} presents detailed statistics of the 355K image-description triplets in our dataset. During the creation of training, validation and test set splits, we ensure absolute exclusivity between patents in the train set and those in the combined validation and test sets, to enable robust out-of-sample evaluation. To achieve this, we randomly sampled $\sim$ 12.6K patents from $\sim$ 60K, representing $\sim$ 82.5K images. From this isolated subset of images, we sample $\sim$17K images each for the val and test set, and discard the remaining images. This sampling technique also helps maintain the diversity within the validation and test sets, thereby providing a fair and representative evaluation. Our detailed descriptions span $\sim$1.7K tokens on average, which is much larger compared to an average token length for popular image captioning benchmarks~\cite{Lin2014mscoco,chen2015coco-caps,sidorov2020textcaps}.

\section{Methodology}
\label{sec:method}


\begin{figure*}[!t]
\centering
\includegraphics[width=\textwidth]{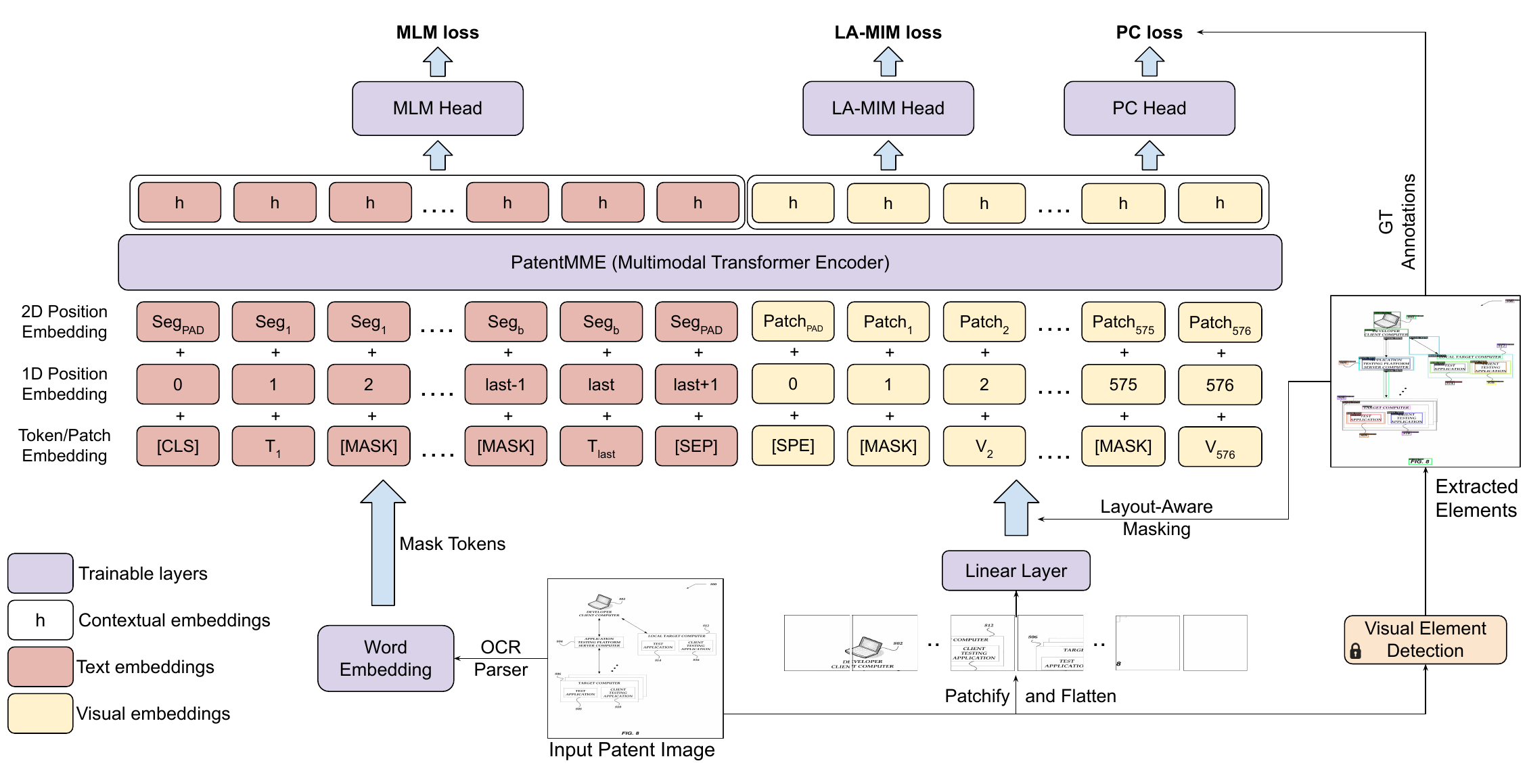}
\caption{\textsc{PatentMME} Architecture. We jointly process OCR tokens and visual embeddings to produce multimodal context-aware embeddings. These contextual embeddings are optimized using our proposed MLM, LA-MIM and PC objectives. } 
\label{fig:patent_mme}
\end{figure*}

Our approach is inspired by the recent success of large multimodal models like MiniGPT-4~\cite{Zhu2023MiniGPT4EV} and LLaVA~\cite{LLaVA, LLaVA-1.5}, which have demonstrated state-of-the-art performance on several benchmarks by effectively aligning visual and textual modalities.
We introduce \model{}, which combines our domain-adapted version of the LLaMA language model, namely \textsc{PatentLLaMA}, with our novel visual encoder specialized for patent figures, namely \textsc{PatentMME}. In this section, we describe the architectures of \textsc{PatentMME} and \textsc{PatentLLaMA}, and the overall framework of \model{}. 

\subsection{\textsc{PatentMME}: Encoder for Patent Figures}
The Vision Transformer (ViT)~\cite{ViT}, commonly used as a vision encoder in existing image captioning frameworks, is typically pre-trained on natural scene images, which are fundamentally different from patent figures. A better suited encoder is perhaps  LayoutLM~\cite{xu2020layoutlm,xu2021layoutlmv2,huang2022layoutlmv3} which has shown impressive performance in document image understanding tasks. However, patent figures have a sparse layout compared to dense document images and are characterized by specific structured visual syntax. Unlike document images, patent figures comprise labeled nodes interconnected with arrows and accompanied by textual elements. The semantic relationship between these diagrammatic constituents is paramount for decoding the inventive concepts and technical specifications elucidated within the patent figures.
We, therefore, build on the existing document image understanding capabilities of LayoutLMv3~\cite{huang2022layoutlmv3} and pre-train it with novel objectives, specifically tailored to capture the structural information of patent figures.
\subsubsection{Architecture:}
The proposed \textsc{PatentMME} shares its architecture with LayoutLMv3~\cite{huang2022layoutlmv3} and is a multi-modal transformer model that processes image, text, and document layout information jointly. The overall architecture of the model is illustrated in Fig.~\ref{fig:patent_mme}. Given an input patent figure $I$, the OCR text is extracted using off-the-shelf Tesseract OCR engine~\cite{tesseractOCR}. The image is then down-scaled to $H \times W$ and split into non-overlapping patches of $p$ dimensions each, resulting in $M=HW/p^2$ image patches. 
The OCR extracted text is tokenized using the BPE tokenizer~\cite{bpe} and represented using a learnable embedding matrix. Following~\cite{huang2022layoutlmv3}, learnable 1D-position embeddings and 2D segment-level layout-position embeddings are added to the word embeddings, resulting in the final text embeddings. 
The image embeddings are created by linear projection of flattened image patches and combining them with learnable 1D position embeddings and 2D spatial embeddings. 
We use images of size $I \in \mathbb{R}^{3 \times 384 \times 384}$, i.e., $H=W=384$. With $p=16$ this results in $M=576$ patches. The higher resolution helps preserve intricate structural details of patent figures, such as node labels and arrows.

\subsubsection{Pre-training data and annotations:}
To enable large-scale in-domain pre-training of \textsc{PatentMME}, we crawled a set of 900K+ patent figures corresponding to the patent IDs from the Harvard USPTO Patent Dataset (HUPD)~\cite{suzgun2024harvard}. For a fair evaluation, appropriate care has been taken to avoid any overlap of the sample with the validation and testing split of our dataset.

For robust patent-figures'-specific pretraining, we define loss functions that leverage patent diagram specific elements like nodes, node labels, figure labels, text and arrows.
To extract such elements, we train a Faster-RCNN~\cite{ren2015faster_rcnn} based visual element detection network on 350 manually annotated patent figures, sampled randomly from our training data. 
The trained model is then used to infer elements from all training images, which is used to provide weak ground-truth labels during \textsc{PatentMME} training.
We show inference samples of this model in Appendix B.

\subsubsection{Pre-training Loss Formulations:}
To enhance the vision encoder's capability in capturing fine-grained structural details of patent figures, we pre-train \textsc{PatentMME} using novel layout-aware masked image modeling (LAMIM) and image patch classification (PC) objectives, along with the established masked language modeling (MLM) loss. We describe these losses in the following text.\\
\textbf{Notation:} We use $R$ and $T$ to denote the set of image patches (regions) and OCR tokens, respectively. Further, $X_{m}$ and $X_{um}$ denote the masked and unmasked parts of the modality $X$. The probability distribution generated by our \textsc{PatentMME} model and the set of categories of visual elements that can be detected by our detection network by $p_\theta$ and $C$, respectively.
\begin{figure}[!t]
\includegraphics[width=\columnwidth]{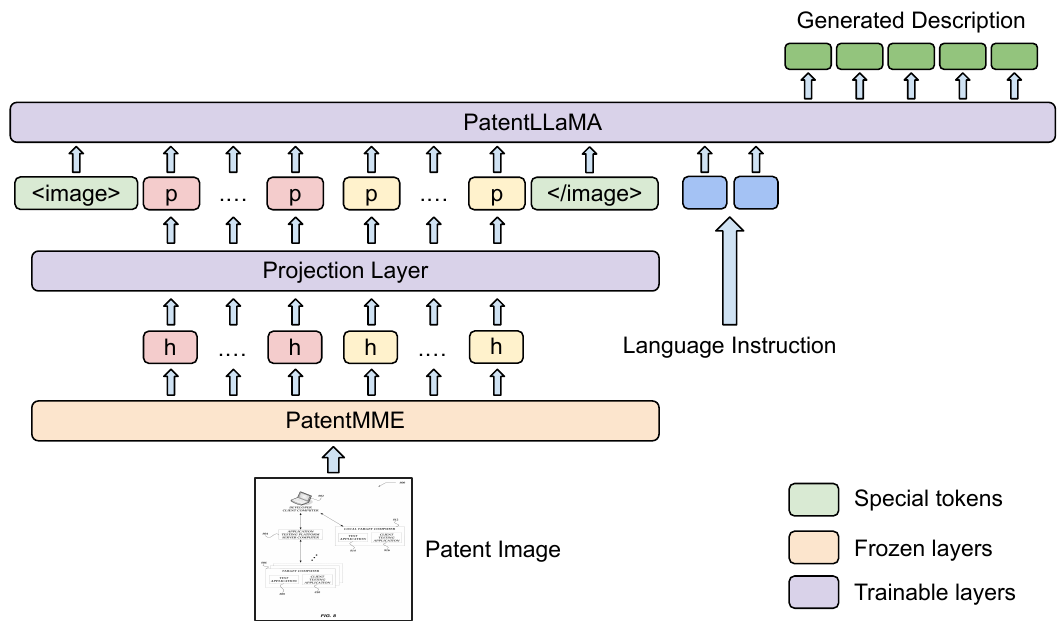}
\caption{\label{fig:patent_lmm}\textsc{PatentLMM} Architecture. Language Instruction is a fixed prompt guiding the model to generate either brief or detailed descriptions.}

\end{figure}


\noindent(i) \textbf{
Masked Language Modeling (MLM)}. Similar to 
LayoutLMv3~\cite{huang2022layoutlmv3},
we randomly mask 30\% of the OCR text tokens and optimize the model to predict the masked tokens, encouraging it to learn patent-specific textual semantics. This loss is computed as follows:
\begin{equation}
\mathcal{L}_{MLM}(\mathbf{\theta}) = -\sum_{i \in T_m} \log p_{\theta}(t_{i} \mid R_{um}, T_{um}),
\end{equation}
where $t_{i}$ denotes the correct masked text tokens.

\noindent(ii) \textbf{
Layout-Aware Masked Image Modeling (LAMIM)}. 
We utilize masked image modeling to learn visual representations by randomly masking 40\% of the image patches. Since the patent figures are more sparse compared to dense document images, we mask only the image patches that contain atleast one of the following five elements: nodes, node labels, figure labels, text and arrows. 
This strategy helps to avoid masking blank regions in the patent figures, and hence learn robust visual representations. Our formulation of the LAMIM objective is similar to BEiT~\cite{bao2022beit} and therefore requires a discrete image tokenizer. 
We choose OCR-VQGAN~\cite{rodriguez2023ocr-vqgan} since its tokenized image representation is capable of handling textual information better than competing works dVAE~\cite{ramesh2021dall.e-dvae} and VQGAN~\cite{Esser_2021taming}. We compute this loss as follows:

\begin{equation}
\mathcal{L}_{MIM}(\theta) = -\sum_{i \in R_m} \log p_{\theta}(r_{i} \mid R_{um}, T_{um}),
\end{equation}
where $p_{i}$ denotes the correct masked image patches.

\noindent(iii) \textbf{
Patch Classification (PC).} In this multi-label binary classification objective, we classify each of the $M$ image patches into one or more of the following five categories: \emph{node}, \emph{node label}, \emph{figure label}, \emph{text}, and \emph{arrows}. This objective which is mathematically computed as follows, helps the model learn discriminative representations for different visual elements 
in patent figures. 
\begin{equation}
    \mathcal{L}_{PC} = -\frac{1}{M} \sum_{i=1}^M \sum_{j=1}^{|C|} \left[ y_{ij} \log(\hat{y}_{ij}) + (1 - y_{ij}) \log(1 - \hat{y}_{ij}) \right],
\end{equation}
where $\hat{y}_{ij}$ denotes the probability of patch $i$ belonging to class $j$, and $y_{ij}$ is the binary ground truth label obtained using the visual element detector.

\subsection{\textsc{PatentLLaMA}: Description Generator}

\textsc{PatentLLaMA} is a domain-adapted version of the LLaMA-2 7B model for the patent domain. We continue to pre-train the LLaMA-2 7B model using LoRA~\cite{hu2022lora} adapters, on the descriptions from HUPD patent dataset~\cite{suzgun2024harvard}, to bias the model to generate the language inherent to patent documents. To avoid any train-test leakage, we ensure that we use the HUPD dataset after removing patent documents corresponding to the validation and test splits of our \dataset{} dataset. 

\subsection{\textsc{PatentLMM}}

Inspired by recent multimodal LLM studies like MiniGPT-4~\cite{Zhu2023MiniGPT4EV} and LLaVA~\cite{LLaVA, LLaVA-1.5}, we integrate \textsc{PatentMME} and \textsc{PatentLLaMA} through a single MLP network to exploit their pre-trained representations. 
The detailed architecture for \textsc{PatentLMM} is illustrated in Fig.~\ref{fig:patent_lmm}. Given a patent figure, we first obtain its layout-aware text and visual representations from frozen \textsc{PatentMME}. 
These representations are projected into the input embedding space of \textsc{PatentLLaMA} using a projection MLP, 
and the \textsc{PatentLLaMA} is finetuned to maximize the likelihood of the corresponding description conditioned on these projected representations. 
\section{Experiments}
\label{sec:experiments}

\begin{table*}[!t]
\centering
\scriptsize
\begin{tabular}{c l |r| r r r r r r r| r r r r r r r r r}
\toprule
\textbf{Setup}&\textbf{Method}&\textbf{\# Parameters}&\textbf{B-2}&\textbf{B-4} & \textbf{Avg. B} & \textbf{R-1} & \textbf{R-2} & \textbf{R-L} & \textbf{M}&\textbf{B-2}&\textbf{B-4} & \textbf{Avg. B} & \textbf{R-1} & \textbf{R-2} & \textbf{R-L} & \textbf{M}\\
\midrule
&&&\multicolumn{7}{c|}{Brief}&\multicolumn{7}{c}{Detailed}\\
\midrule
\multirow{5}{*}{\rotatebox{90}{Zero-shot}} &
BLIP-2 & 2.7B& 1.01 & 0.03 & 1.62 & 15.43 & 1.70 & 12.47 & 6.72& 0.00 &0.00 & 0.01 & 3.24 & 0.49 & 2.84 & 1.00\\
&TextMonkey &9.8B& 0.91 & 0.11 & 1.12 & 13.00 & 4.60 & 12.16 & 7.14 & 0.10 & 0.03 & 0.10 & 6.18 & 2.38 & 4.83 & 2.64\\
&PEGASUS &568M& 3.18 & 0.13 & 4.20 & 14.68 & 2.33 & 11.46 & 13.24& 0.86 & 0.04 & 1.12 & 12.26 & 1.96 & 9.47 &  5.18\\
&mPLUG-owl2& 7.5B&3.64 & 0.36 & 4.47 & 21.47 & 5.10 & 19.41 & 13.07 & 3.65 & 0.49 & 3.40 & 23.75 & 5.39 & 14.85 &  11.83\\
&UReader& 7.2B & 3.54 & 0.35 & 4.50 & 20.90 & 4.56 & 17.85 & 13.45 & 0.05 & 0.01 & 0.06 & 5.15 & 1.54 & 4.49 & 2.04\\
&LLaVA-1.5&7.4B& 4.52 & 0.24 & 4.71 & 17.59 & 3.27 & 14.63 & 15.74& 3.65 & 0.37 & 3.36 & 23.75 & 4.63 & 14.71 & 11.69\\
&GPT-4V& Unknown&20.74 & 8.56 & 18.68 & 36.07 & 15.65 & 31.89 & 32.88& 19.61 & 6.05 & 18.26 & 39.95 & 12.14 & 20.16 & 27.31\\
\midrule
\multirow{7}{*}{\rotatebox{90}{Finetuned}}
&Pegasus &568M& 2.44 & 0.14 & 4.03 & 13.86 & 1.55 & 11.52 & 11.62& 5.80 & 0.41 & 6.33 & 19.28 & 2.24 & 15.27  & 12.11\\
&GIT & 681M&26.95 & 15.33 & 24.78 & 45.28 & 27.17 & 42.29 & 44.27& 6.33 & 1.18 & 6.23 & 13.66 & 3.17 & 10.87 & 10.68\\
&BLIP& 252M&24.62 & 12.52 & 22.40 & 42.59 & 23.78 & 39.16 & 42.84& 5.45 & 1.05 & 5.31 & 12.42 & 2.89 & 9.46 & 9.55\\
&MiniGPT-4&7.8B (3.2M) & 30.57 & 17.96 & 28.13 & 43.53 & 25.33 & 40.35 & 43.03 & 11.01 & 2.81 & 10.26 & 28.91 & 6.23 & 15.67 & 16.65\\
&OFA& 472M & 33.01 & 21.76 & 31.24 & 54.26 & 37.94 & 51.47 & 44.89& 15.76 & 7.23 & 14.93 & 33.20 & 13.70 & 22.89 & 21.17 \\
&LLaVA-1.5& 7.4B (341M) &36.64 & 25.00 & 34.37 & 48.92 & 32.01 & 45.87  & 48.23& 20.90 & 11.12 & 19.81 & 36.86 & 15.68 & 24.48 & 24.71 \\
&\textbf{\textsc{PatentLMM}} &7.4B (341M)& \textbf{46.40} & \textbf{36.66} & \textbf{44.59} & \textbf{56.68} & \textbf{42.63} & \textbf{54.18}  & \textbf{56.44}& \textbf{25.42} & \textbf{15.02} & \textbf{24.24} & \textbf{40.70} & \textbf{19.27} & \textbf{27.54} &  \textbf{28.39} \\
\bottomrule
\end{tabular}
\caption{Quantitative results on \dataset{} (test set) for brief and detailed description generation (B=BLEU, R=ROUGE, M=METEOR). Number in parenthesis under {\scriptsize \# \textbf{Parameters}} column denote number of trainable parameters.}
\label{tab:MainResults}
\end{table*}


\subsection{Experimental Setup}

\subsubsection{\textsc{PatentMME}:} \textsc{PatentMME} is initialized with LayoutLMv3-Large to inherit its document understanding capabilities. For each of the three losses discussed in Section~\ref{sec:method}, the text and image embeddings obtained from \textsc{PatentMME} are projected through separate MLPs (loss heads) before the loss is calculated. Since the network weights already have a good initialization, to prevent major changes in weights of the multimodal transformer, we adopt two-step training. During Step-1, the weights of the multimodal transformer remain frozen and only the loss heads are trained for 1 epoch with a higher learning rate of 1e-3 and 1K warm-up steps to learn good initialization. During Step 2, the entire model is trained end-to-end for 8 epochs with a lower learning rate of 5e-5 and with 10K warm-up steps. The \textsc{PatentMME} model is trained on 8$\times$V100 GPUs, with an effective batch size of 64 and Adam~\cite{kingma2014adam} optimizer.

\subsubsection{\textsc{PatentLMM}:}  Following the standard practice~\cite{LLaVA-1.5}, we train our \textsc{PatentLMM} model in two stages. To align the patent figure representations obtained from \textsc{PatentMME} with the input latent space of \textsc{PatentLLaMA}, we train only the projection layer in the first stage, keeping all other parameters frozen. During stage 2, we add LoRA adapters to all the linear layers of the PatentLLaMA module, except for the language modeling head, whose weights remain frozen. The weights of \textsc{PatentMME} are kept frozen throughout. We train our \textsc{PatentLMM} with an effective batch size of 192 on 3$\times$A100 GPUs (40 GB). Stage 1 training progresses at a higher learning rate of 1e-3, and stage 2 training takes place at a learning rate of 2e-4 with a cosine schedule, for 12K steps using Adam optimizer. We train separate LMMs for brief and detailed descriptions.

Overall, training \textsc{PatentLMM} is a three-phase process. Firstly, we train the \textsc{PatentMME} encoder in a semi-supervised fashion by leveraging a vast amount of patent figures corresponding to patents in the HUPD dataset. Secondly, we domain-adapt the LLaMA-2 7B model on the HUPD patent text data to create \textsc{PatentLLaMA}. Lastly, we integrate \textsc{PatentMME} and \textsc{PatentLLaMA} to create \textsc{PatentLMM}, and train it following the two-stage process.

\subsection{Baselines}
We benchmark the performance of various baselines on our proposed \dataset{} dataset in the zero-shot and fine-tuned setup. 
We benchmark the text-only baseline Pegasus~\cite{Zhang2019PEGASUSPW} by generating patent figure descriptions from OCR tokens extracted from patent figures. For image captioning baselines, we study the state-of-the-art models GIT~\cite{Wang2022GITAG}, BLIP~\cite{Li2022BLIPBL} and OFA~\cite{OFA}. We further compare our method with recent multimodal LLMs such as UReader~\cite{ye2023ureader}, TextMonkey~\cite{liu2024textmonkey}, mPLUG-owl2~\cite{ye2023mplug-owl2}, BLIP-2~\cite{Li2023BLIP2BL}, MiniGPT-4~\cite{Zhu2023MiniGPT4EV}, LLaVA-1.5~\cite{LLaVA-1.5} and the closed GPT-4V model~\cite{achiam2023gpt4}
. GPT-4V prompt is listed in Appendix C.3.

To measure the description generation performance of these models, we use standard image captioning metrics such as BLEU~\cite{bleu}, ROUGE~\cite{lin-2004-rouge} and METEOR~\cite{banerjee-lavie-2005-meteor}. Higher values for all the scores are desired. A detailed description of these metrics is provided in Appendix C.1.


\subsection{Results and Discussion}
The quantitative performance comparison for the brief and detailed description generation task is reported in Table~\ref{tab:MainResults}. In the zero-shot setting, GPT-4V demonstrated superior performance among baselines across all metrics, significantly outperforming other baselines owing to its large scale and the diverse data it has seen during its pre-training. The poor zero-shot performance of other baselines highlights the gap in their pre-training data and the nature of patent figures and descriptions. The fine-tuned models outperform their zero-shot counterparts, highlighting the importance of task-specific training for these models. MiniGPT-4 and LLaVA-1.5 utilize a frozen pre-trained ViT trained on web-scale natural images, which results in suboptimal representation of patent figures. Similarly, OFA also enforces these priors by utilizing a pre-trained discrete image tokenizer. On the other hand, \model{} gives a boost of $\sim$ 8\% across all metrics, signifying the importance of better domain knowledge embedded in it through the proposed \textsc{PatentMME} pretraining and PatentLLaMA.

Similar to brief description generation, GPT-4V outperformed all other baselines for the detailed description generation task in the zero-shot setting. We observe that majority of the baselines struggle with performance in the zero-shot setup. In the fine-tuned setting, our \model{} maintained its superior performance, achieving the highest scores across all metrics. This consistent top performance for both brief and detailed descriptions suggests the efficacy of our proposed approach for the task of generating descriptions from patent figures. The overall lower scores for detailed descriptions can be attributed to their comprehensiveness, complexity, and length, requiring models to capture and generate more nuanced and detailed information.






\subsubsection{Ablations:}

We perform the following three ablation studies to quantify the impact of different components of our proposed \textsc{PatentLMM} model:

\noindent\textbf{(i) \textsc{PatentMME} Pre-training objectives}: Table~\ref{tab:ablationPatentMME} shows the ablation results with combinations of pre-training objectives for the brief description generation. We observe that using a combination of MLM and LAMIM leads to better results compared to the pre-trained LayoutLMv3. Further, the PC loss also improves the performance of the model, when pre-trained with HUPD images data. A similar ablation for detailed descriptions is reported in Appendix C.2.


\begin{table}[!t]
\centering
\scriptsize
\begin{tabular}{l r r r r r r r}
\toprule
\textbf{Pre-training}&\textbf{B-2}&\textbf{B-4} & \textbf{Avg. B} & \textbf{R-1} & \textbf{R-2} & \textbf{R-L} & \textbf{M}\\
\midrule
  Pretrained LayoutLMv3 & 42.81 & 32.50 & 40.86 & 53.68 & 38.88 & 51.07 & 53.34 \\  
  w/ MLM + LAMIM & 45.24 & 35.33 & 43.39 & 55.69 & 41.38 & 53.20 & 55.34\\
  w/ MLM+LAMIM+PC & \textbf{46.39} & \textbf{36.65} & \textbf{44.59} & \textbf{56.68} & \textbf{42.62} & \textbf{54.18} & \textbf{56.44}\\
\bottomrule
\end{tabular}
\caption{Ablation study to quantify the impact of pre-training objectives of \textsc{PatentMME} on the overall performance of \textsc{PatentLMM} on brief descriptions generation task. All models are trained with \textsc{PatentLLaMA}.}
\label{tab:ablationPatentMME}
\end{table}

\noindent\textbf{(ii) Importance of OCR tokens}: In this ablation, we study whether avoiding passing OCR tokens to \textsc{PatentLMM} causes any drop in the performance of brief description generation. 
We experiment with two ablations: (1) OCR tokens are used for \textsc{PatentMME} pretraining but not for \textsc{PatentLMM} training, and (2) OCR tokens are used for \textsc{PatentMME} pretraining and for \textsc{PatentLMM} training but not at inference time. Table~\ref{tab:ablationOCRTokens} shows that it is important to use OCR tokens in the entire pipeline for the best results.

\begin{table}[!t]
\centering
\scriptsize
\begin{tabular}{p{0.12\columnwidth} p{0.135\columnwidth} r r r r r r r}
\toprule
\textbf{OCR in training?}&\textbf{OCR in Inference?}&\textbf{B-2}&\textbf{B-4} & \textbf{Avg. B} & \textbf{R-1} & \textbf{R-2} & \textbf{R-L} & \textbf{M}\\
\midrule
  No & No &30.32&19.17&28.30&41.61&25.21&38.95&41.46 \\
  Yes & No &11.51&2.77 &9.83&24.38&7.92&21.68&22.52 \\
  Yes&Yes & \textbf{46.40} & \textbf{36.66} & \textbf{44.59} & \textbf{56.68} & \textbf{42.63} & \textbf{54.18}  & \textbf{56.44} \\
\bottomrule
\end{tabular}
\caption{Ablation study to quantify the importance of OCR tokens on the overall performance of \textsc{PatentLMM} on brief descriptions generation task.} 
\label{tab:ablationOCRTokens}
\end{table}

\noindent\textbf{(iii) PatentLMM Training:} We report an additional ablation study in Appendix C.2 to quantify the advantage of using PatentLLaMA against the pre-trained LLaMA model.

\subsubsection{Qualitative Analysis:}

Fig.~\ref{fig:qual_res} shows an example brief and detailed description generated by \model{} for a test sample. The generated brief description, more specifically, terms the electronic device shown in the image as a mobile smartphone. The generated detailed description provides a comprehensive overview of the electronic device 104, its components, and their functions. It covers most of the key elements mentioned in the ground truth, including the processor 114, the memory 116, the display 118, and the user interface 122. However, there are some omissions, like Graphics Component 154 and Communication Component 150. More case studies are provided in Appendix D.


\noindent\textbf{Error analysis}: 
We perform a thorough manual error analysis on a set of 50 samples drawn from our test set to identify some prominent errors in the descriptions generated by our \model{} model. 
We identify five main error categories as follows.
(i) Hallucination in figure labeling occurs in 3 brief and 3 detailed descriptions.
(ii) Hallucination in 4 brief descriptions and 7 detailed descriptions was due to little or no OCR detectable text in the figures.
(iii) Incorrect association of node labels occurs when the wiggly arrows connecting node labels to respective nodes are misinterpreted or ignored due to downsampling of the image before being passed to \textsc{PatentMME}. This was observed in 10 detailed descriptions. 
(iv) A similar misinterpretation due to downsampling is often the cause of hallucinated node labels in 12 detailed descriptions. 
(v) 
Cross-figure references in the descriptions establish the interconnection between various aspects and provide a complete picture of the presented technical invention. The figures may be related hierarchically (systems vs components), sequentially (steps of a process), different views (top-bottom-left-right), or in other ways. Since we train \model{} to generate descriptions for individual patent figures, our model hallucinates the cross-figure references for 2 brief and 5 detailed descriptions. Qualitative examples are presented in Appendix D.2.


\noindent\textbf{GPT-4V Evaluation Results}:  Apart from small-scale manual error analysis, we utilize the GPT-4V model to qualitatively evaluate the performance of LLaVA-1.5 and our proposed \model{} model on the brief and detailed description generation task for a set of 1000 samples. We input the GPT-4V model with the patent figure, the ground truth description and the description generated using these models, along with the special instruction prompt. The instruction prompt instructs the GPT-4V model to rate the generated description on the following criterion: Relevance, Accuracy, Completeness (with respect to input image), Fluency and Coverage (with respect to input image and ground truth description) on an integer scale of 0 to 2. To mitigate randomness in scores, we set the temperature parameter to 0 for the GPT-4V model and created five versions of the instruction prompt. The scores obtained from each of the prompts for each criterion are then averaged. Table~\ref{tab:gtp4_qual_eval} shows that our system generates high-quality results.

\begin{table}[!t]
\centering
\scriptsize
\begin{tabular}{l rr rr rr r}
\toprule
Description&Method&Rel.&Acc.&Compl.&Coh.&Fluency&Cover.\\
\midrule
\multirow{2}{*}{Brief}&LLaVA-1.5&1.38&1.06&1.01&1.85&1.98&0.98\\
&Ours&1.44&1.18&1.17&1.91&2.00&1.15\\
\multirow{2}{*}{Detailed}&LLaVA-1.5&0.75&0.75&0.73&1.07&1.69&0.71\\
&Ours&0.90&0.78&0.76&1.15&1.85&0.75\\
\bottomrule
\end{tabular}
\caption{GPT-4V evaluation on a set of 1K samples.} 
\label{tab:gtp4_qual_eval}
\end{table}

\section{Conclusion and Future Work}
\label{sec:conclusion_future-work}
Our work addresses the existing gap in the automated generation of patent figure descriptions by introducing \dataset{}, a comprehensive dataset of patent figures and their corresponding brief and detailed descriptions. We further proposed \textsc{PatentLMM}, a large multimodal model comprising a domain-specialized image encoder \textsc{PatentMME} and a domain-adapted patentLLaMA model for generating brief and detailed descriptions from patent figures. Extensive experiments demonstrated that our proposed \model{} outperforms competent baselines by significant margins. 
Future research in this direction can explore experiments with patents in multiple languages, patent document-level reasoning to allow for cross-figure references while generating descriptions, incorporating external knowledge bases from technical domains to improve the performance of detailed description generation, and generation of grounded descriptions.

\section*{Acknowledgements}
This work was supported by the Microsoft Academic Partnership Grant (MAPG) 2023.

\bibliography{references}
\clearpage
\appendix
\section*{Appendix}

\begin{figure*}[!t]
\includegraphics[width=\textwidth]{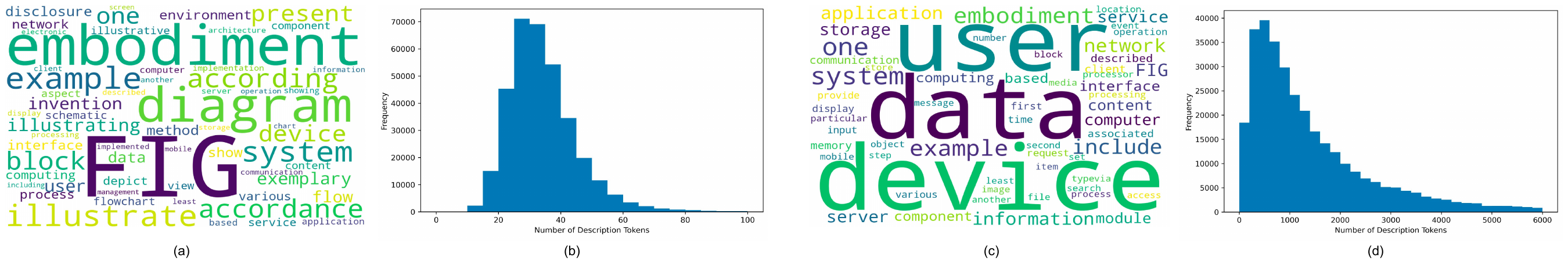}
\caption{\dataset{} Analysis. (a) and (c): Word clouds of most common occurrences in brief and detailed descriptions respectively. (b) and (d):  frequency distribution of description lengths for the brief and detailed descriptions respectively.}
\label{fig:dataset_analysis}
\end{figure*} 

\begin{figure*}[t!]
\includegraphics[width=\textwidth]{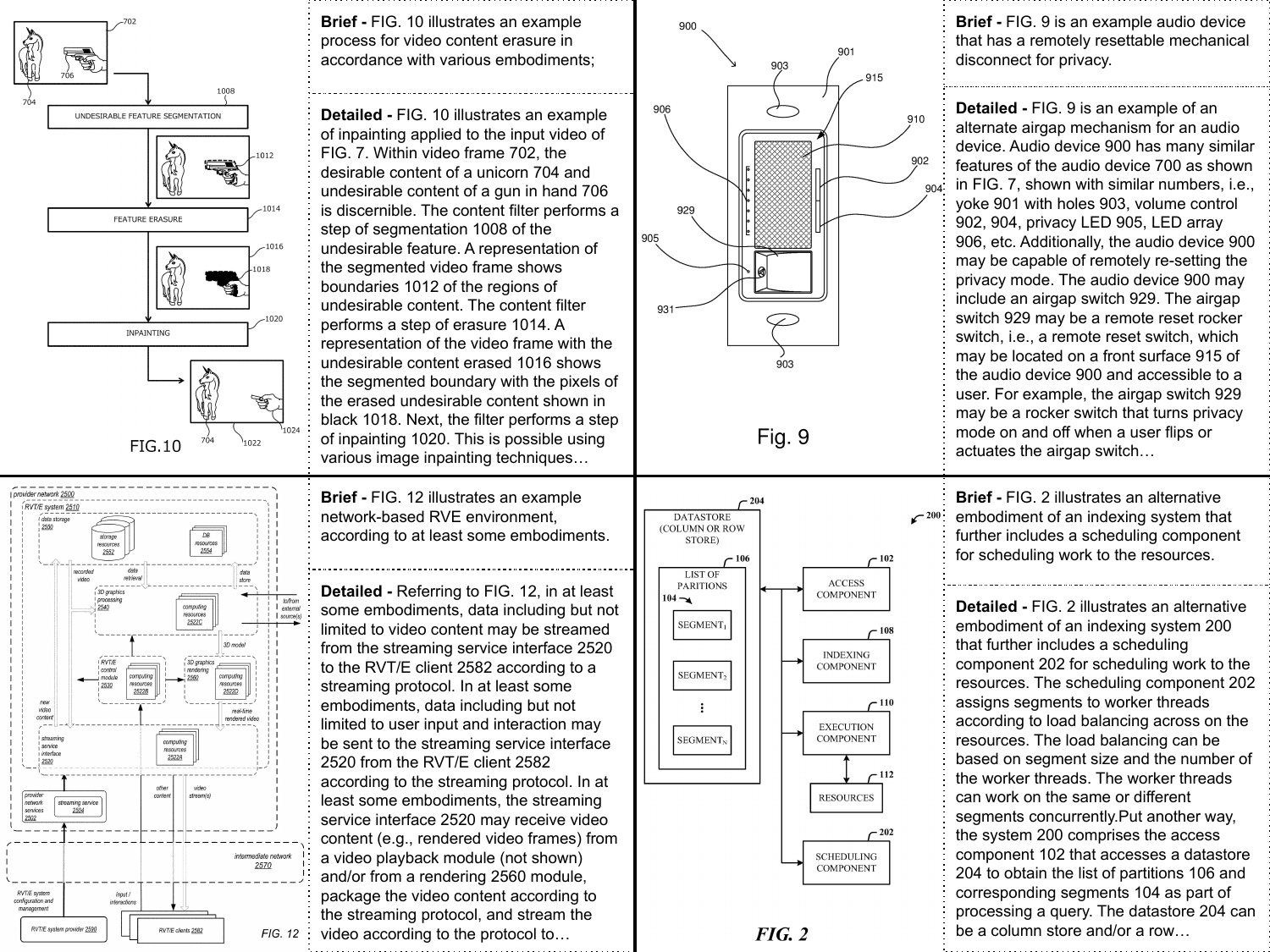}
\caption{Samples from our \dataset{} dataset showing $\langle$patent figure, brief description, detailed description$\rangle$ triplets.}
\label{fig:dataset_samples}
\end{figure*} 

\section{\dataset{}}
\subsection{Dataset Curation}
To create a comprehensive dataset, we crawled a diverse set of over 90K US patent documents published between 1900 and January 2023. This involved searching for various companies on Google Patents and downloading respective CSV files, with detailed relevant patent metadata like ID, assignee, publication date, patent URLs, etc. We use the patent-ids and URLs to download the HTML documents. 
We parsed these HTML documents to extract the image URLs, and downloaded $\sim$900K images, ensuring a rich and representative corpus of technical illustrations. In the following sections, we describe the preprocessing and description extraction stages: 
 
\subsubsection{Pre-processing of Patent Images}
We manually analyzed a random set of 500 patent images from our collection, and encountered considerable noise attributed to the vast diversity. To alleviate this noise, we implemented a series of filtering steps as follows.

\begin{enumerate}
    \item \textbf{Correcting Image Orientation}: We identified that $\sim$40\% of the analysed images were vertically oriented. To rectify this automatically, we compared the average length of OCR tokens extracted using PaddleOCR~\cite{ppocr-v2} for the original image and the 90$^{\circ}$-rotated image, and saved the image with greater average OCR length.
    
    \item \textbf{Redundancy Removal}: We eliminated the first occurrence of representative figure images, which were repeated twice for each patent. 
    
    \item \textbf{Discarding Multi-Figure Images}: Around 7\% of the analysed images had multiple figures per image. To maintain a focus on singular representative visuals per image, we extract figure labels using PaddleOCR~\cite{ppocr-v2}. Then, we remove a small proportion of images containing multiple occurrences of figure labels.
    
    \item \textbf{Graph/Plot/Chart Removal}: Around 5\% images in our analysed data depicted graphical plots. Prior works~\cite{tang2023vistext,linecap,masry2022chartqa,liu2023matcha} have studied their captioning in detail through specialized handling, so we discard these images by training a ResNet-50-based binary classifier on a subset of 300 manually annotated images, achieving a 98\% validation accuracy.
    

    \item \textbf{Publication Date Filtering}: We observed a specific convention in HTML tags for patents published after 2004. So, to ensure consistency in HTML tags for easier description extraction, we discarded patents published before 2005. This resulted in our final set of patents published from Jan 2005 to Jan 2023.
\end{enumerate}

After these image-based pre-processing steps, we end up with a final set of $\sim$429K images corresponding to $\sim$64K unique patents. 

\subsubsection{Extracting Descriptions of Patent Figures}
We obtain the brief and detailed descriptions for each image from the corresponding patent HTML document as follows. We first extract figure labels from the images utilizing PaddleOCR~\cite{ppocr-v2}. Next, we extract the content within the \textit{brief-description-of-drawings} tag in the HTML. Within this tag, there is a child \textit{description-line} or \textit{description-paragraph} tag corresponding to every image. Hence, for each image, using its figure label, we extract the text enclosed within the corresponding \textit{description-line} or \textit{description-paragraph} tag as its brief description.

For detailed descriptions, we consider the paragraphs falling after the \textit{brief-description-of-drawings} tag. For each new \textit{description-line}/\textit{description-paragraph} tag, we check if its first sentence contains the \textit{figref} tag. If yes, the tag text is attributed to the referred figure, else we append the tag text to the previously referred figures' description. In rare cases, if the OCR extracts incorrect figure labels, we cannot obtain descriptions for such images and hence we exclude such images from our dataset. Overall, this leads to our final dataset with $\sim$355K images spanning $\sim$60K patent documents.

\subsection{Quality Assessment for the Proposed \dataset{}}
To assess the quality of our automatic description extraction heuristics, we manually annotated a random set of 100 patent images with their brief and detailed descriptions and computed the sentence-level precision and recall of the extracted descriptions against the ground-truth descriptions. For brief descriptions, both precision and recall scores were 100\%. This is expected since brief descriptions span single tags, making their rule-based extraction and OCR-based matching almost error-free. For detailed descriptions, the extracted descriptions match with the ground truth with a precision of 90.81\% and a recall of 91.96\%. Through manual analysis, we identified two primary reasons for the lower scores on detailed descriptions: (a) for the last referred figure in the HTML, sometimes the description includes concluding paragraphs not specifically relevant to the figure, reducing the precision of the descriptions for such figures; (b) some sentences in the description contain references to multiple figures. In this paper, we study the problem of generating descriptions for individual figures. Hence, we discard such sentences, leading to slightly lower recall for those figures. Finally, to establish a consistent evaluation framework across baselines with varying context windows, we clip the detailed descriptions to 500 tokens. 
\begin{figure*}[!t]
\centering
\includegraphics[width=0.98\textwidth]{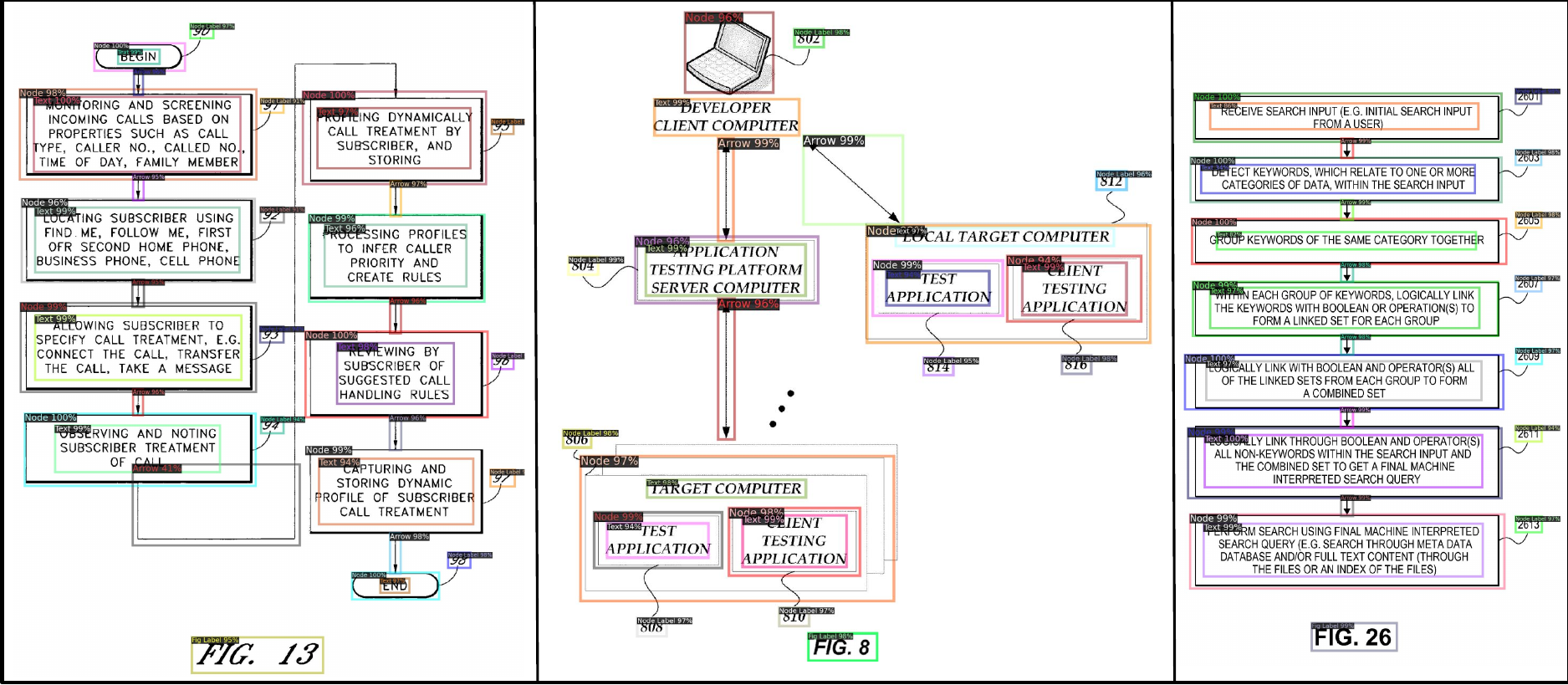}
\caption{Annotations of patent image elements obtained using our visual element detection model on a selection of test samples.}
\label{fig:frcnn_sample}
\end{figure*}

Overall, the quality of the proposed dataset is robust, with high precision and recall scores for brief descriptions and slightly lower but still strong scores for detailed descriptions.

\subsection{Dataset Analysis and Examples}
Fig.~\ref{fig:dataset_analysis}(a) and (c) illustrate the word clouds of the most frequent words in brief and detailed descriptions, respectively. Further, Fig.~\ref{fig:dataset_analysis}(b) and (d) show the frequency distribution of description lengths for the brief and detailed descriptions, respectively.

Fig.~\ref{fig:dataset_samples} shows a selection of example samples -- $\langle$patent figure, brief description, detailed description$\rangle$ from our \dataset{} dataset.

\section{Details on patent Visual Element Detector}
We manually annotated 400 patent images sampled randomly from our training data. The annotations were in the form of bounding boxes for the following five categories: nodes, node labels, text, arrows, and figure labels. We split this dataset into a training split consisting of 350 samples and a test split consisting of 50 samples. We then use the training split to fine-tune a F-RCNN~\cite{ren2015faster_rcnn} model with ResNet101~\cite{he2016resnet} backbone, pre-trained on the MS-COCO dataset. We finetune this model with a learning rate of 1e-4 until convergence. This helps us obtain an AP@50 score of 92.52, and an AP@75 score of 64.34 on the test set. We show a few examples of the annotations obtained using the trained network in Fig.~\ref{fig:frcnn_sample}.

\section{Additional Experiments and Details}
\subsection{Evaluation Metrics}
To measure the description generation performance, we use the standard image captioning metrics such as BLEU~\cite{bleu}, ROUGE~\cite{lin-2004-rouge} and METEOR~\cite{banerjee-lavie-2005-meteor}. Higher values for all the scores are desired. 

\noindent \textbf{BLEU-n}~\cite{bleu} calculates the n-gram overlap between the generated and reference texts, taking into account precision with which captions are generated, and a brevity penalty for shorter texts. We report BLEU-2, BLEU-4 and Avg. BLEU for all our experiments. The Avg. Blue score averages BLEU-1, BLEU-2, BLEU-3 and BLEU-4 metric.  \\
\noindent \textbf{ROUGE}~\cite{lin-2004-rouge} also measures the overlap between the generated and reference texts. 
ROUGE-N measures N-gram overlap while ROUGE-L measures the overlap based on the longest common subsequence between the generated and reference texts. We report ROUGE-1, ROUGE-2 and ROUGE-L for our experiments. \\
\noindent \textbf{METEOR}~\cite{banerjee-lavie-2005-meteor} is computed based on the explicit word-to-word matches between the generated and reference texts. It considers not only exact word matches but also stem, synonym, and paraphrase matches, as well as applies weighted penalties for incorrect word order.

\subsection{Additional Ablations}

\subsubsection{\textsc{PatentLMM} Training:}  Table~\ref{tab:ablationPatentLMM} shows the ablation results with different design choices for training of the decoder LLM of \textsc{PatentLMM} for the brief description generation task. We show results without stage 2 training of the \textsc{PatentLMM} (rows 1 and 3). We also show results when the decoder LLM is initialized using LLaMA-2 versus \textsc{PatentLLaMA}. We observe that stage-1 training is clearly not enough and the decoder LLM needs to be finetuned for the patent description task to generate reasonable descriptions. We also observe that using \textsc{PatentLLaMA} leads to significantly better results compared to just using the standard LLaMA-2 model. This shows that domain adaptive pretraining with HUPD patent text is important for better performance.

\begin{table}[!t]
\centering
\scriptsize
\caption{Ablation study to quantify the impact of the decoder LLM on the overall performance of \textsc{PatentLMM} on brief descriptions generation task. We report the results of \textsc{PatentLMM} with PatentMME vision encoder, at both stages of training.}
\label{tab:ablationPatentLMM}
\begin{tabular}{c l r r r r r r r}
\toprule
\textbf{Stage}&\textbf{LLM Init.}&\textbf{B-2}&\textbf{B-4} & \textbf{Avg. B} & \textbf{R-1} & \textbf{R-2} & \textbf{R-L} & \textbf{M}\\
\midrule
  1 & LLaMA-2 & 1.06 & 0.02 & 1.58 & 6.70 & 0.86 & 5.84 & 10.04 \\
  2 & LLaMA-2 & 43.54 & 33.50 & 41.66 & 54.35 & 39.76 & 51.81 & 53.82 \\
  1&\textsc{PatentLLaMA} & 1.08 & 0.02&1.61 &7.08& 0.87 &5.99& 10.32\\
  2&\textsc{PatentLLaMA} & \textbf{46.39} & \textbf{36.65} & \textbf{44.59} & \textbf{56.68} & \textbf{42.62} & \textbf{54.18} & \textbf{56.44}\\
\bottomrule
\end{tabular}
\end{table}

\subsubsection{Pre-training ablation of PatentMME for detailed description:} In Table~\ref{tab:ablationPatentMMEDetailed}, we observe that using a combination of MLM, LAMIM and PC losses leads to better results compared to the pretrained LayoutLMv3. Thus, domain specific pre-training with HUPD image data is helpful even for detailed description generation.

\begin{table}[!t]
\centering
\scriptsize
\caption{Ablation study to quantify the impact of pre-training objectives of \textsc{PatentMME} on the overall performance of \textsc{PatentLMM} on detailed descriptions generation task. All models are trained with \textsc{PatentLLaMA}.
}
\label{tab:ablationPatentMMEDetailed}
\begin{tabular}{l r r r r r r r}
\toprule
\textbf{Pre-training}&\textbf{B-2}&\textbf{B-4} & \textbf{Avg. B} & \textbf{R-1} & \textbf{R-2} & \textbf{R-L} & \textbf{M}\\
\midrule
  Pretrained LayoutLMv3 & 23.84 & 13.84 & 22.73 & 39.28 & 18.06 & 26.44 & 27.04 \\  
  w/ MLM+LAMIM+PC & \textbf{25.42} & \textbf{15.02} & \textbf{24.24} & \textbf{40.70} & \textbf{19.27} & \textbf{27.54} &  \textbf{28.39} \\
\bottomrule
\end{tabular}
\end{table}

\subsection{Evaluation of GPT-4V as a baseline}
We utilize the following prompt to evaluate the patent description generation capabilities of the GPT-4V model as one of the baselines in the zero-shot setting.

\noindent \texttt{\textbf{System Prompt:} You have been hired to draft patents for the world's largest companies. Your primary task is to help your company in drafting patent documents.\\
\textbf{User Prompt:} You will be provided with a figure which will be part of a new patent that your company is planning to file.
Your task is to generate a brief description for the patent figure and it will be used as a part of the patent document, and will be included under the `BRIEF DESCRIPTION OF THE DRAWINGS' section of the patent document.\\ \\
IMAGE:$\#$image\_url$\#$\\ \\
Output the brief desctiption in $<$results$>$$<$/results$>$ tag.
}

\noindent Please note that the provided prompt generates brief descriptions. For detailed descriptions, we simply replace all occurrences of ``brief'' with ``detailed'' in the prompt.

\subsection{Using GPT-4V for qualitative evaluation}
For evaluating the quality of the brief and detailed descriptions generated by our \model{} model, we utilize GPT-4V as an evaluator. More specifically, we provide the GPT-4V model with the patent image, the ground truth description, and the description generated using our model. These inputs are accompanied by a prompt that instructs the GPT-4V model to rate the generated description against the ground truth by giving it an integer score from {0, 1, 2}. We use the following prompt to accomplish this:

\noindent \texttt{\textbf{System Prompt:} You are a helpful legal assistant that has been hired to quantitatively evaluate the quality of brief descriptions of patent figures generated from a black-box system, against the original brief description and the image of the patent figure.
}

\noindent \texttt{\textbf{User Prompt:} You are acting as a critical assistant to a patent examiner. We have developed a model that creates both detailed and brief descriptions from images, and we seek your expertise to evaluate the quality of these generated outputs. Your evaluation should focus on the following criteria:\\
Relevance: The degree to which the description corresponds with the content of the image.\\
Accuracy: The correctness of the specific details provided in the description.\\
Completeness: The extent to which the description addresses all significant elements of the image.\\
Coherence: The logical consistency, clarity, and readability of the description.\\
Fluency: The grammatical and stylistic quality of the text, ensuring it reads smoothly.\\
Coverage: Whether the generated description adequately covers the essential concepts from the ground truth text or image.\\ \\
You will be given the image, the reference ground truth description, and the description generated by the model. Please provide an integer score of 0, 1, or 2 for each criterion, with 0 indicating worse performance, 1 indicating reasonable, and 2 indicating the perfect score.\\ \\
Keep in mind that while the generated description may not exactly match the ground truth, it should still faithfully represent the content of the image. Pay close attention to the patent figure when making your assessments.\\ \\
Ensure that your output is formatted as follows:\\
Relevance: $<$your score$>$\\
Accuracy: $<$your score$>$\\
Completeness: $<$your score$>$\\
Coherence: $<$your score$>$\\
Fluency: $<$your score$>$\\
Coverage: $<$your score$>$\\ \\
IMAGE:\#img\_url\#\\
GROUND TRUTH:\#gt\_desc\#\\
GENERATED DESCRIPTION:\#gen\_desc\#\\ \\
Output the scores for each metric in the prescribed format in $<$results$>$$<$/results$>$ tag.
}

We replace all the instances of ``brief'' with ``detailed'' in the System Prompt to evaluate the detailed descriptions generated by our \model{}.

\section{Additional Qualitative Analysis}
\subsection{Case Studies}
\begin{figure*}
\includegraphics[width=\textwidth]{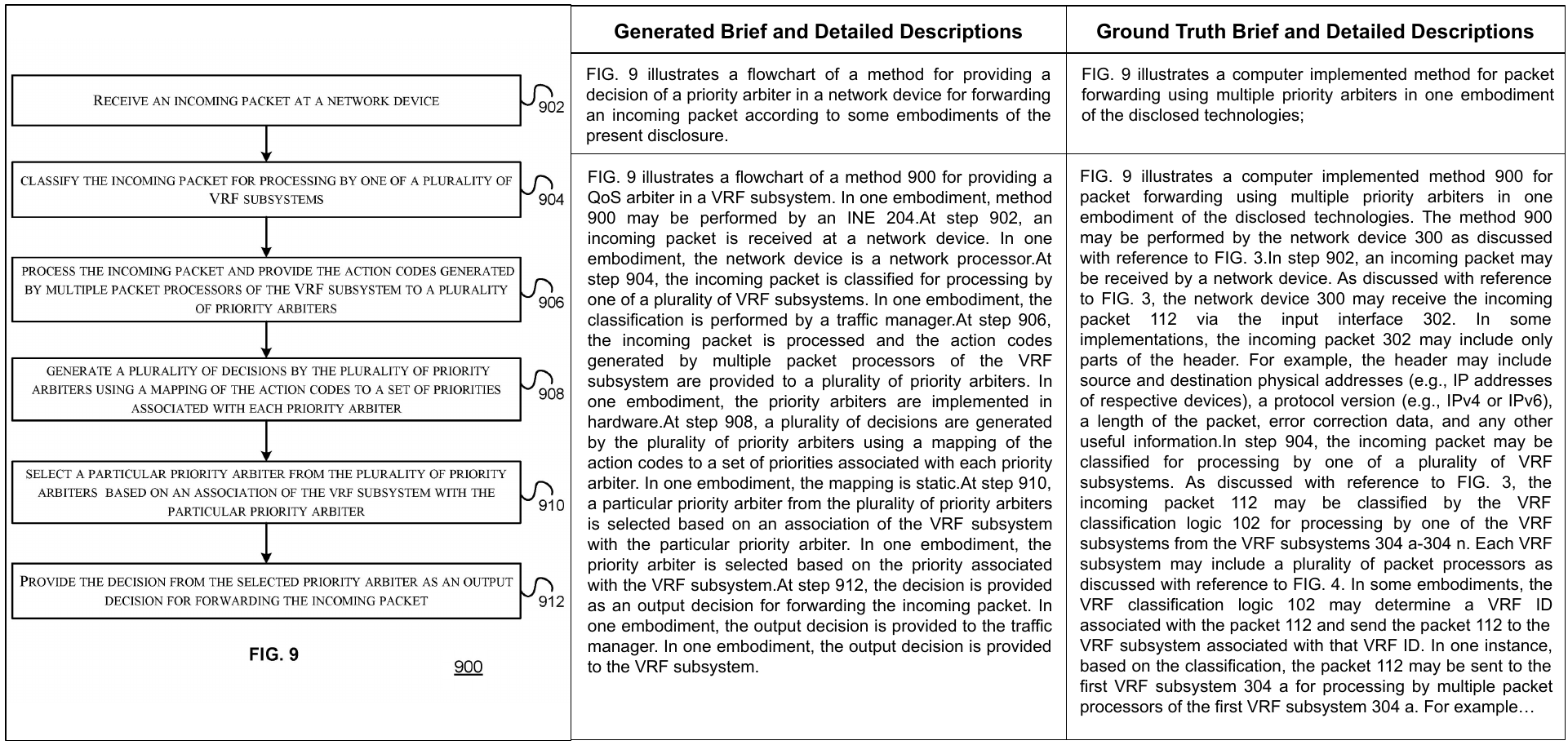}
\caption{Case Study 1: An example of generated and ground truth brief and detailed descriptions using our proposed \model{}.
}
\label{fig:case_study1}
\end{figure*}

We perform rigorous case studies by carefully going through randomly chosen 8 patent images and their corresponding brief and detailed descriptions generated by our approach, critically comparing them against respective ground truths. We summarize our observations for all these examples in the following text:

In Fig.~\ref{fig:case_study1}, the generated brief description accurately describes the flowchart as a method for providing the decision of a priority arbiter in a network device for forwarding an incoming packet. The generated detailed description provides a comprehensive breakdown of the flowchart steps, such as receiving an incoming packet at a network device, classifying the packet for processing by VRF subsystems, processing the packet and generating action codes, generating decisions using priority arbiters, selecting a particular priority arbiter and providing a decision for forwarding the packet. Although the steps and their purposes are described accurately in the generated description, however, when compared to the ground truth detailed description, the generated description is missing reference to Figs. 3 and 4 (cross-figure references) of the same patent. This is because our \model{} is trained to describe each patent image independently, limiting the model's ability to draw connections between related images within the same patent document.

\begin{figure*}
\includegraphics[width=\textwidth]{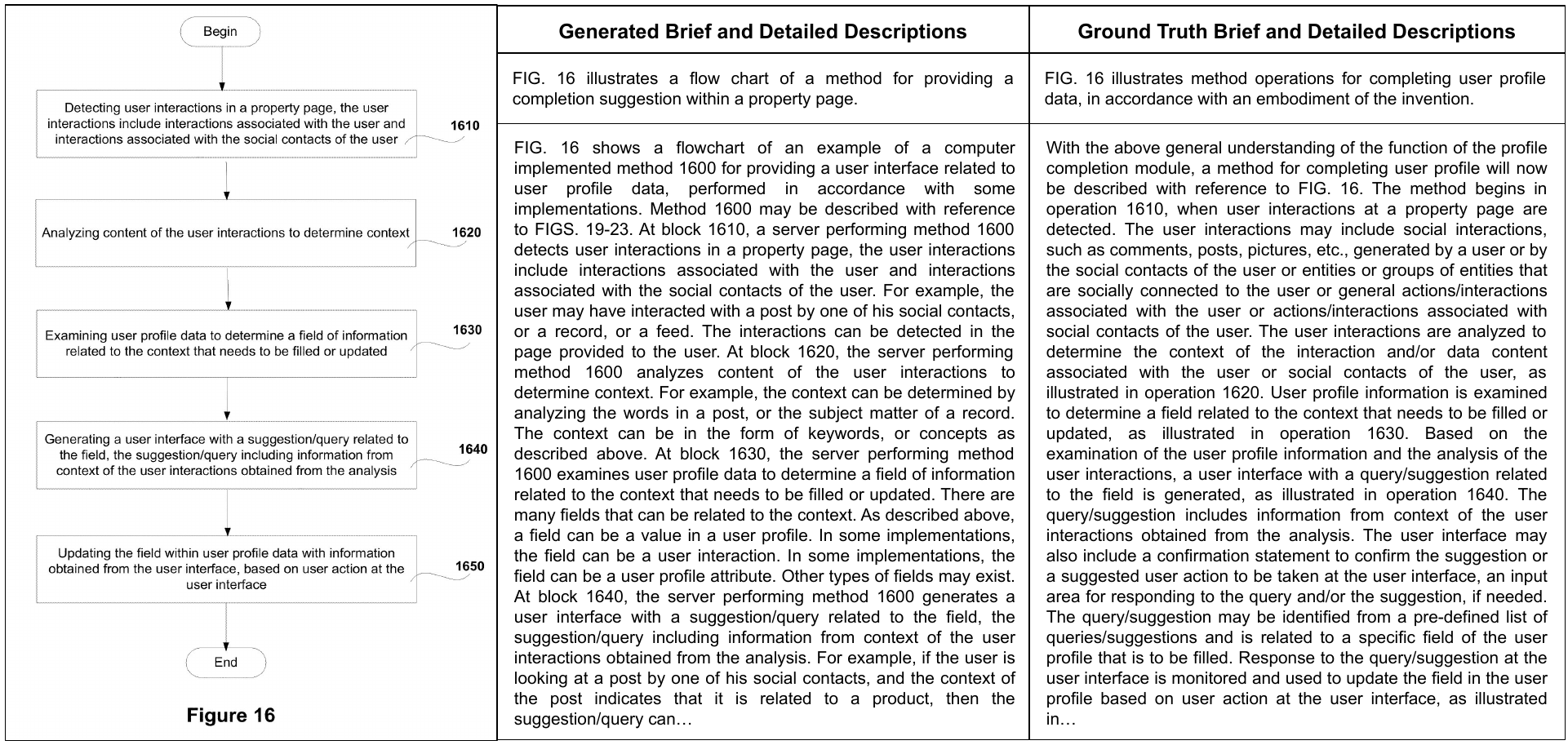}
\caption{Case Study 2: An example of generated and ground truth brief and detailed descriptions using our proposed \model{}.}
\label{fig:case_study2}
\end{figure*}

In Fig.~\ref{fig:case_study2}, the generated brief description is more specific compared to the ground truth by mentioning 'property page', but loses its overall meaning by not talking about 'user profile data'. The generated detailed description provides a comprehensive explanation of the flowchart, covering all the steps shown. It accurately describes the process of detecting user interactions, analyzing content, examining user profile data, generating a user interface with suggestions, and updating the user profile. When compared, the generated and ground truth descriptions mostly differ in their language and level of details only.

\begin{figure*}
\includegraphics[width=\textwidth]{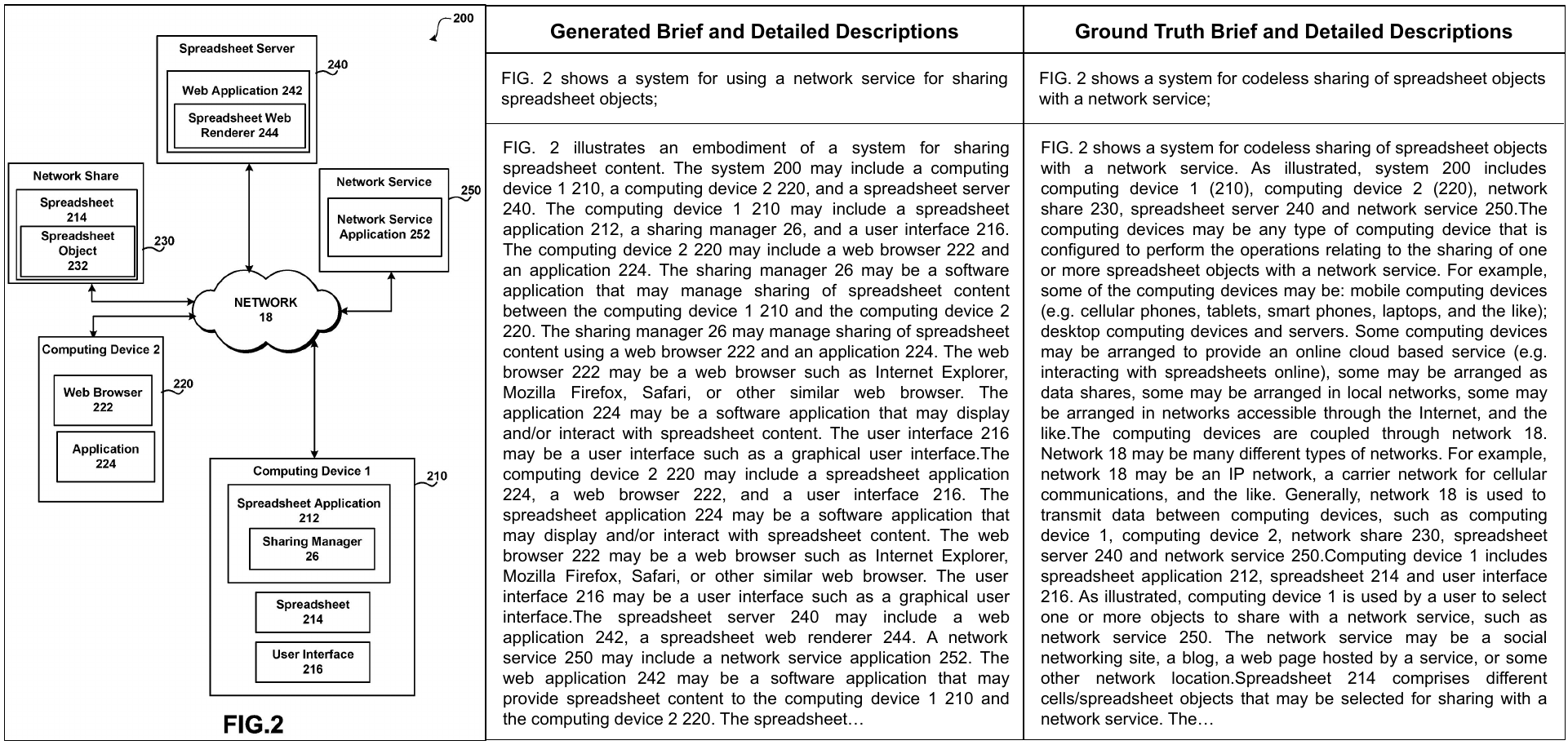}
\caption{Case Study 3: An example of generated and ground truth brief and detailed descriptions using our proposed \model{}.}
\label{fig:case_study3}
\end{figure*}

In Fig.~\ref{fig:case_study3}, the generated brief description focuses on using a network service for sharing spreadsheet objects, which aligns well with the ground truth. The generated detailed description provides a fairly comprehensive overview of the system's components and their interactions like the sharing manager 26, the web browser 222, the application 224, the user interface 216, etc. However, it does not provide as much context on the types of computing devices and network configurations that can be used in the system, unlike the ground truth. Moreover, the term 'codeless sharing', a concept central to the ground truth, is omitted in the generated descriptions.

\begin{figure*}
\includegraphics[width=\textwidth]{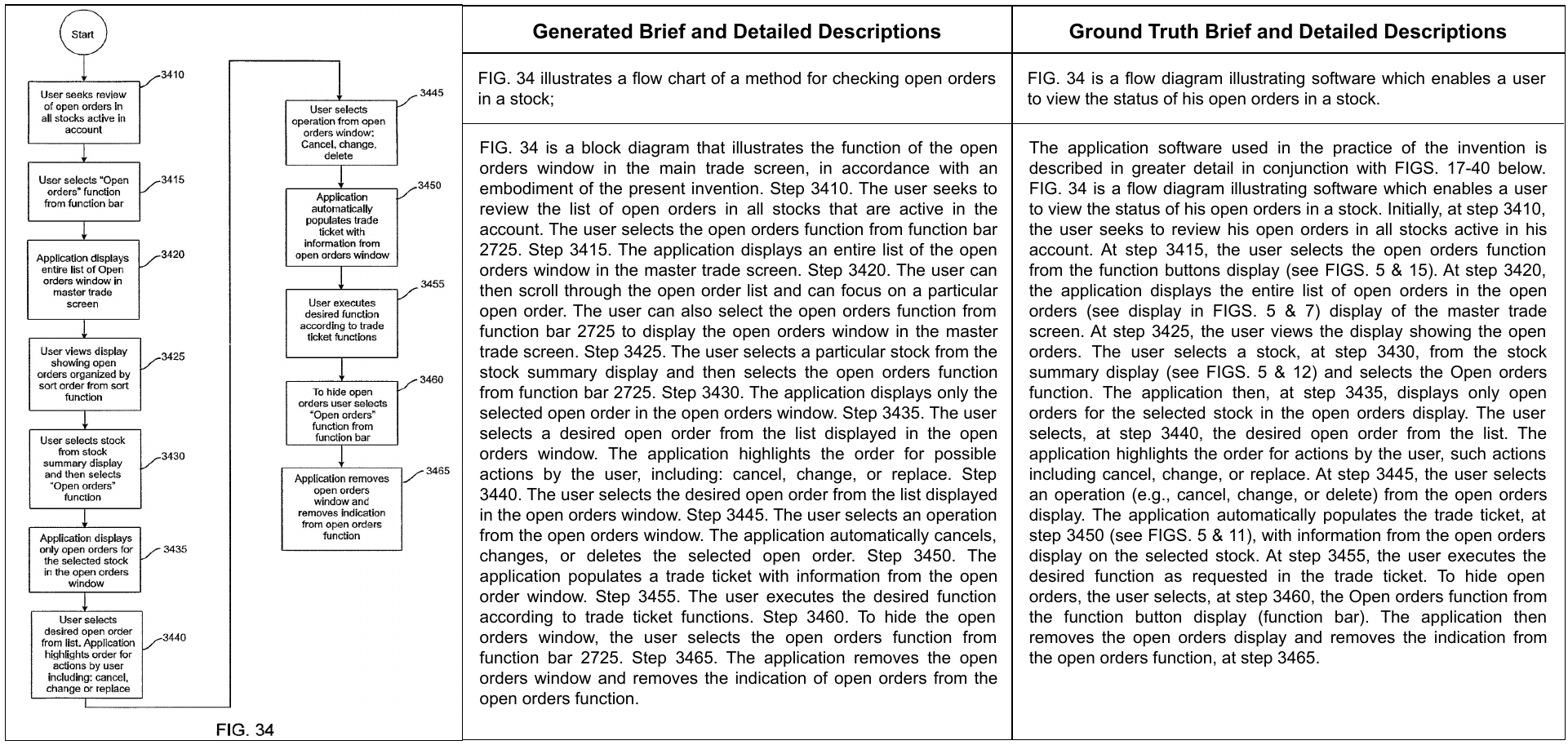}
\caption{Case Study 4: An example of generated and ground truth brief and detailed descriptions using our proposed \model{}.}
\label{fig:case_study4}
\end{figure*}

In Fig.~\ref{fig:case_study4}, the generated brief description correctly describes the flowchart as a method for checking open orders in a stock, which aligns well with the ground truth's emphasis on viewing open order status. The generated detailed description demonstrates a comprehensive understanding of the process flow, accurately describing key steps such as selecting open orders function, displaying the list of open orders, selecting a particular stock, highlighting orders for possible actions, selecting operations like cancel, change, or replace, and populating a trade ticket with information. When compared to the ground truth, some cross-figure references are absent in the generated description. It is however commendable how our \model{} successfully captured the unusual flow indicated by arrows and node labels in this figure despite the downsampling of the image before being passed to the model.

\begin{figure*}
\includegraphics[width=\textwidth]{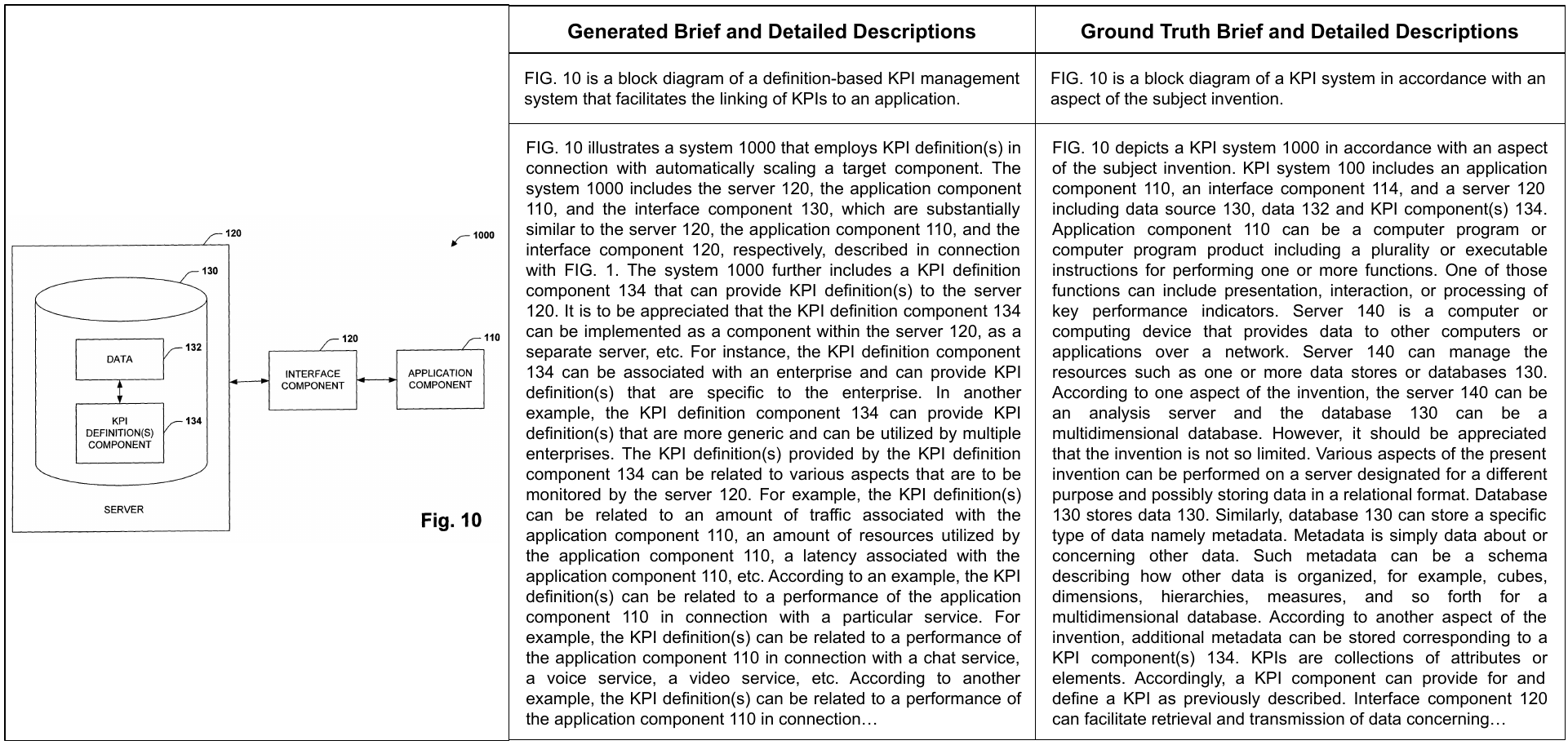}
\caption{Case Study 5: An example of generated and ground truth brief and detailed descriptions using our proposed \model{}.}
\label{fig:case_study5}
\end{figure*}

In Fig.~\ref{fig:case_study5}, the generated brief description accurately identifies the block diagram as a KPI management system. The generated detailed description further captures many key elements correctly. These include the server 120, the application component 110, the interface component 120 and the KPI definition component 134. The generated description however does not mention the data source 130 and the data 132 components shown in the patent figure. Moreover, the description of the capabilities of the server 120 and the nature of the databases 130 is less detailed in the generated description compared to the ground truth.

\begin{figure*}
\includegraphics[width=\textwidth]{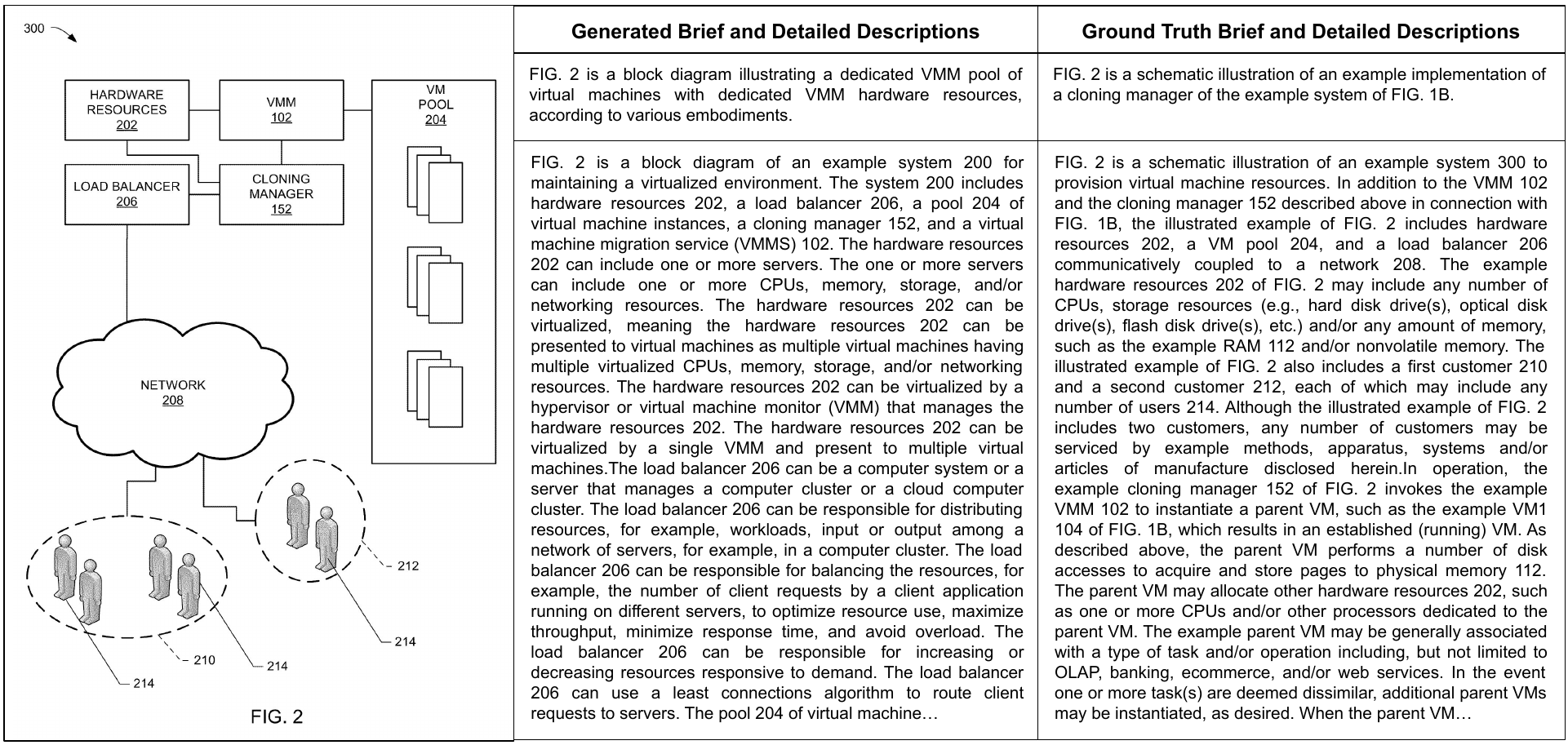}
\caption{Case Study 6: An example of generated and ground truth brief and detailed descriptions using our proposed \model{}.}
\label{fig:case_study6}
\end{figure*}

In Fig.~\ref{fig:case_study6}, the generated brief description correctly identifies the image as a block diagram illustrating a VMM (Virtual Machine Manager) pool with dedicated hardware resources. The generated detailed description captures many system components accurately including the hardware resources 202, the load balancer 206, the cloning manager 152, the VM pool 204 and the VMM 102. While our model accurately describes these components in detail, it incorrectly identifies the system as 200 instead of 300. This misidentification likely stems from the model observing other node labels in the diagram that start with 2 (such as 202, 204, 206, 208, 214) and erroneously extending this pattern to the overall system number. 

\begin{figure*}
\includegraphics[width=\textwidth]{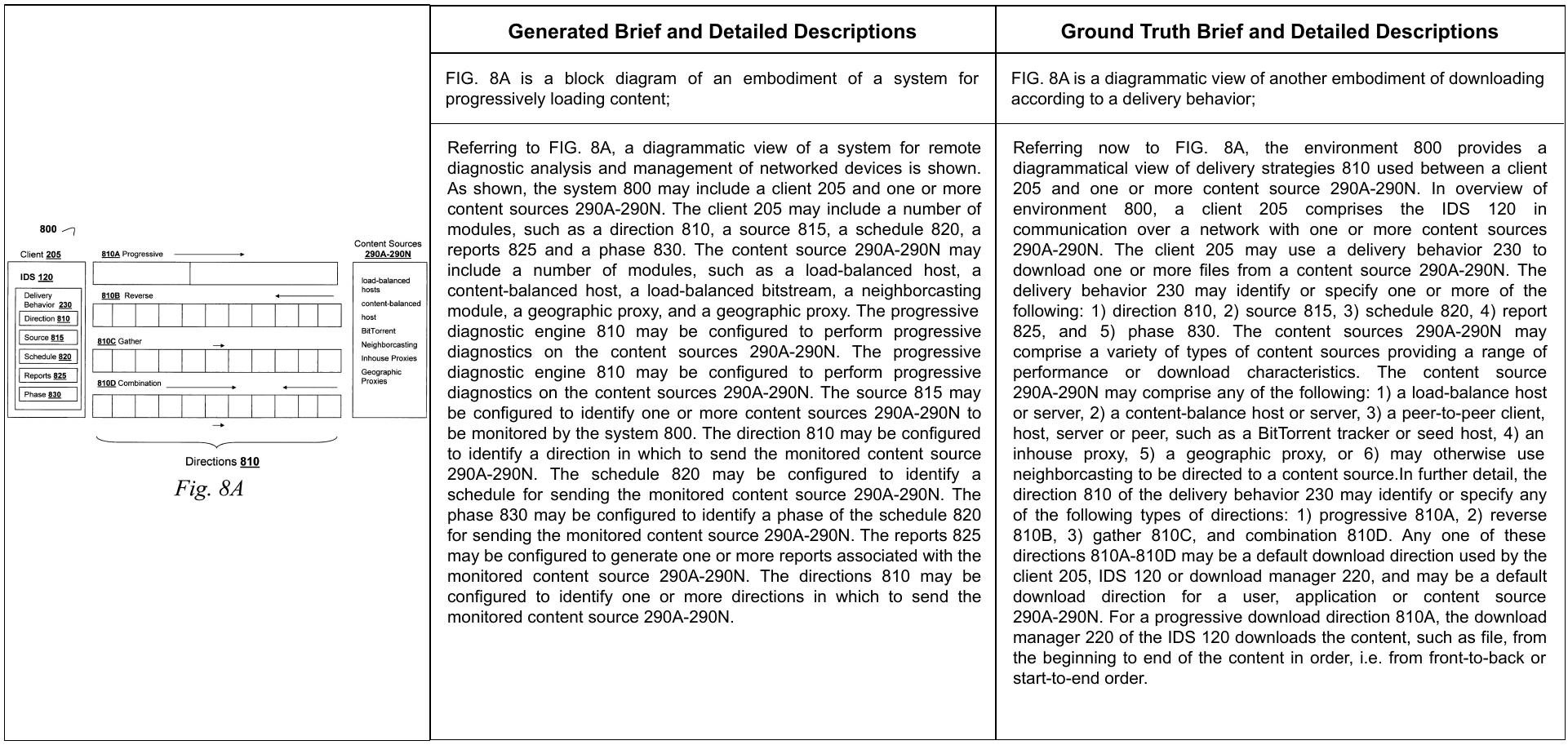}
\caption{Case Study 7: An example of generated and ground truth brief and detailed descriptions using our proposed \model{}.}
\label{fig:case_study7}
\end{figure*}

In Fig.~\ref{fig:case_study7}, the generated brief description accurately captures the general idea of a content delivery system. The generated detailed description correctly identified many components like the client 205, the content sources 290A-290N, direction 810, source 815, schedule 820, reports 825, and phase 830. The description further elaborates on some of these components. However, the generated description focused more on the configuration of various components for monitoring and diagnostics, while the ground truth emphasized on download behaviors and content delivery methods.

\begin{figure*}
\includegraphics[width=\textwidth]{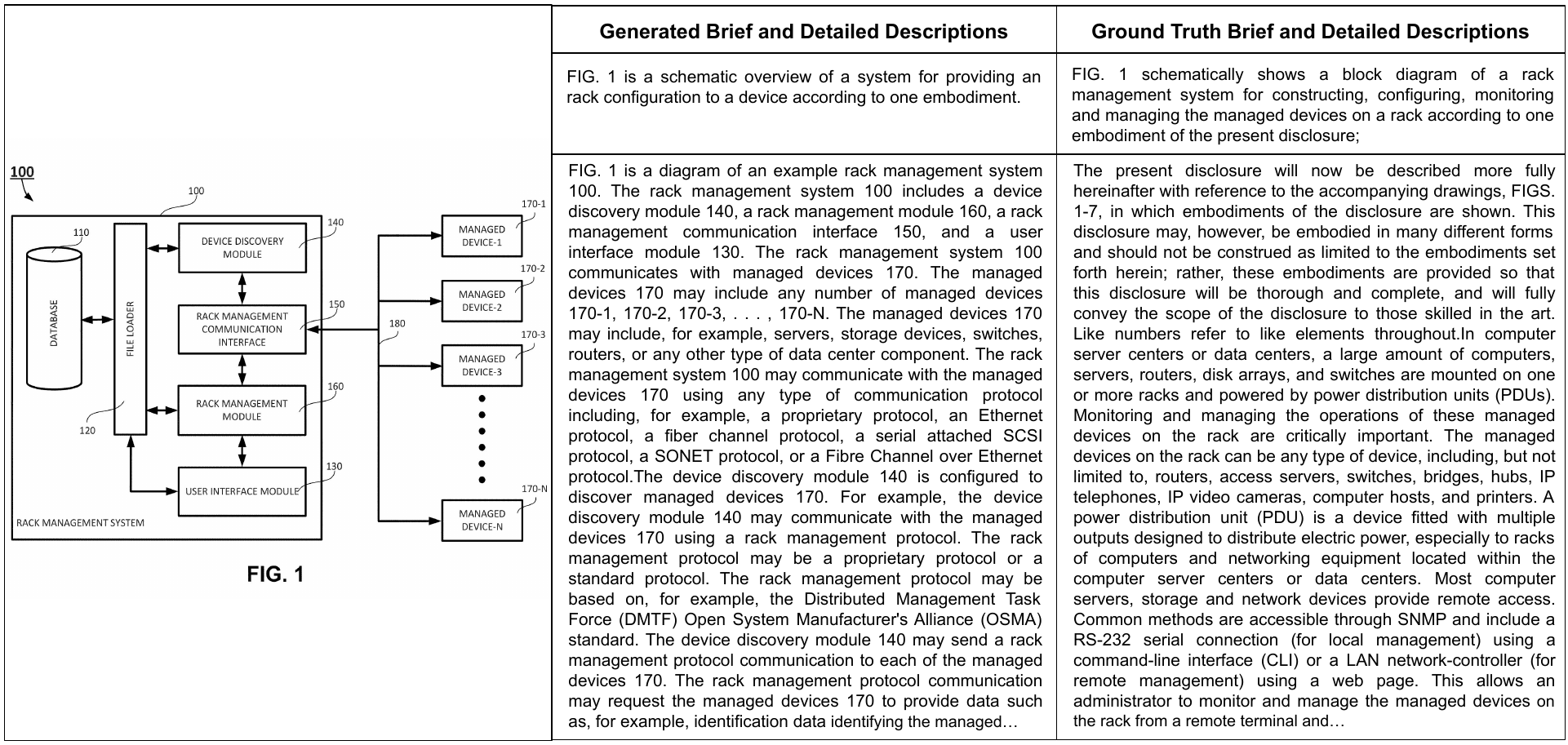}
\caption{Case Study 8: An example of generated and ground truth brief and detailed descriptions using our proposed \model{}.}
\label{fig:case_study8}
\end{figure*}

In Fig.~\ref{fig:case_study8}, the generated brief description, though less comprehensive compared to ground truth, correctly identifies the image as a schematic overview of a system for providing rack configuration to a device. The generated detailed description accurately lists the main components of the system such as the device discovery module 140, the rack management module 160, the rack management communication interface (150), the user interface module (130) and the examples of managed devices 170 (servers, switches, routers). When compared, the generated and ground truth descriptions mostly differ in their level of details for certain components.

\subsection{Failure Cases}
\begin{figure*}
\includegraphics[width=\textwidth]{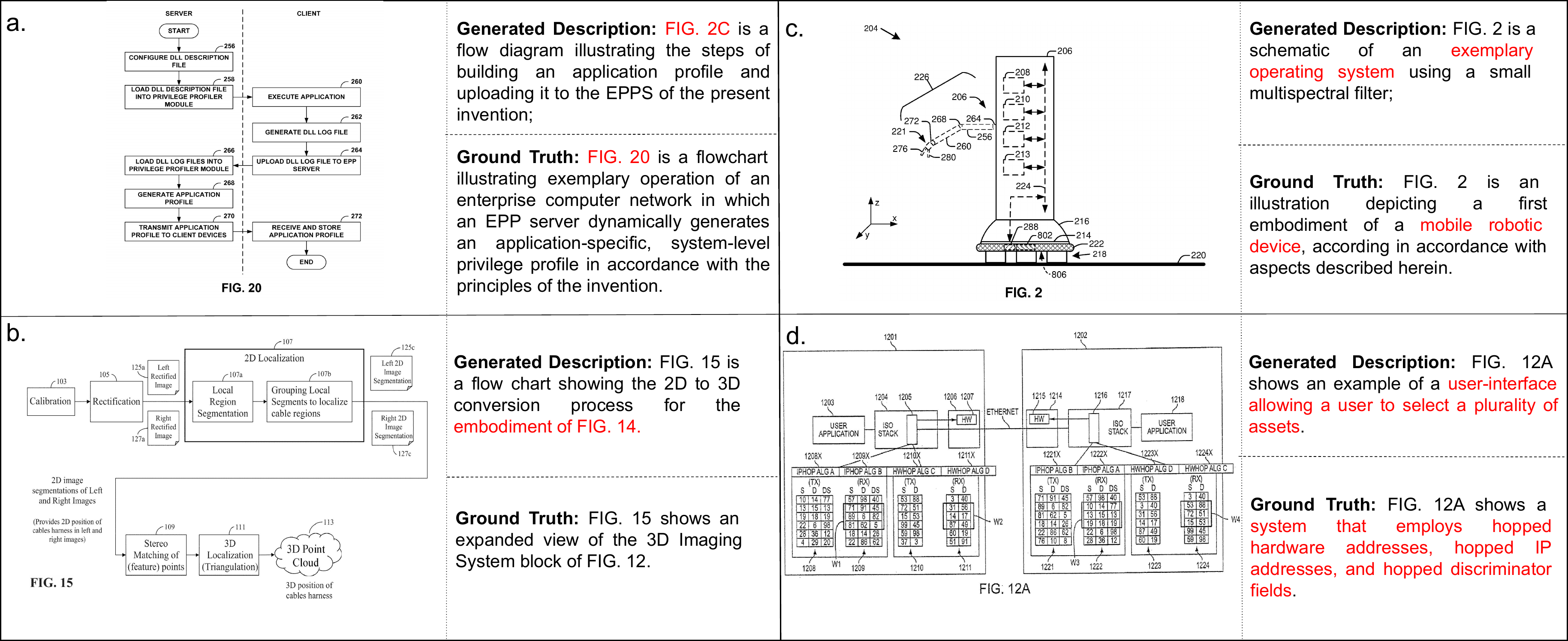}
\caption{Failure Cases for the brief descriptions generated by \model{}.}
\label{fig:brief_error_analysis}
\end{figure*}

Fig.~\ref{fig:brief_error_analysis} demonstrates a few failure cases that occur in the brief descriptions generated by our \model{} model. Specifically, Fig.~\ref{fig:brief_error_analysis}(a) corresponds to the case when the model confuses the OCR (reads Fig 20 as 2C) and generates the figure label incorrectly, Fig.~\ref{fig:brief_error_analysis}(b) demonstrates a case when the model hallucinates reference to another figure of the same patent (also discussed for detailed description for case study 1 (Fig.~\ref{fig:case_study1}) in section D.1), and Figs.~\ref{fig:brief_error_analysis}(c) and (d) demonstrate the case when the model hallucinates the description due to negligible OCR-detectable text in the figure.

\begin{figure*}
\includegraphics[width=\textwidth]{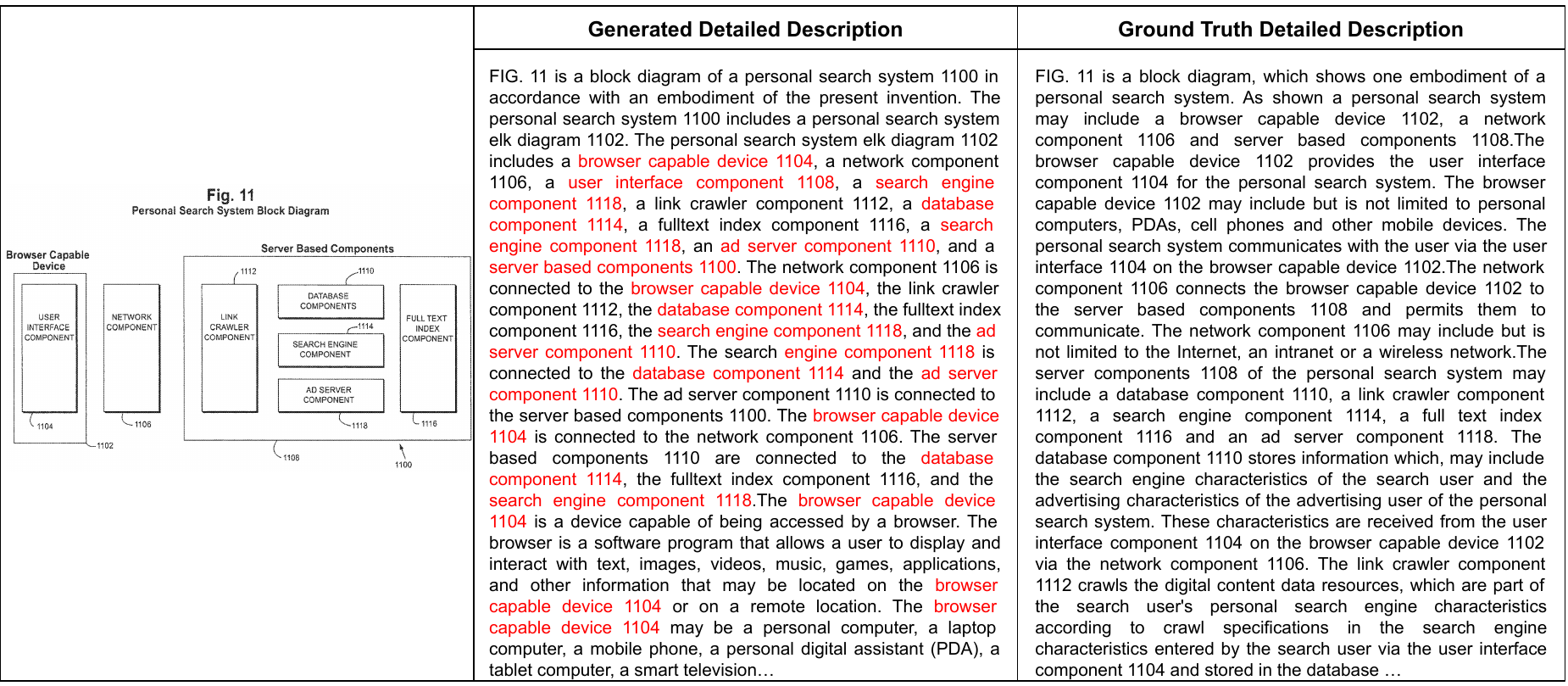}
\caption{Failure Case for the detailed description generated by \model{}. The text in red highlights the incorrect association of nodes and node labels in the generated description.}
\label{fig:detailed_failure_case2}
\end{figure*}

\begin{figure*}
\includegraphics[width=\textwidth]{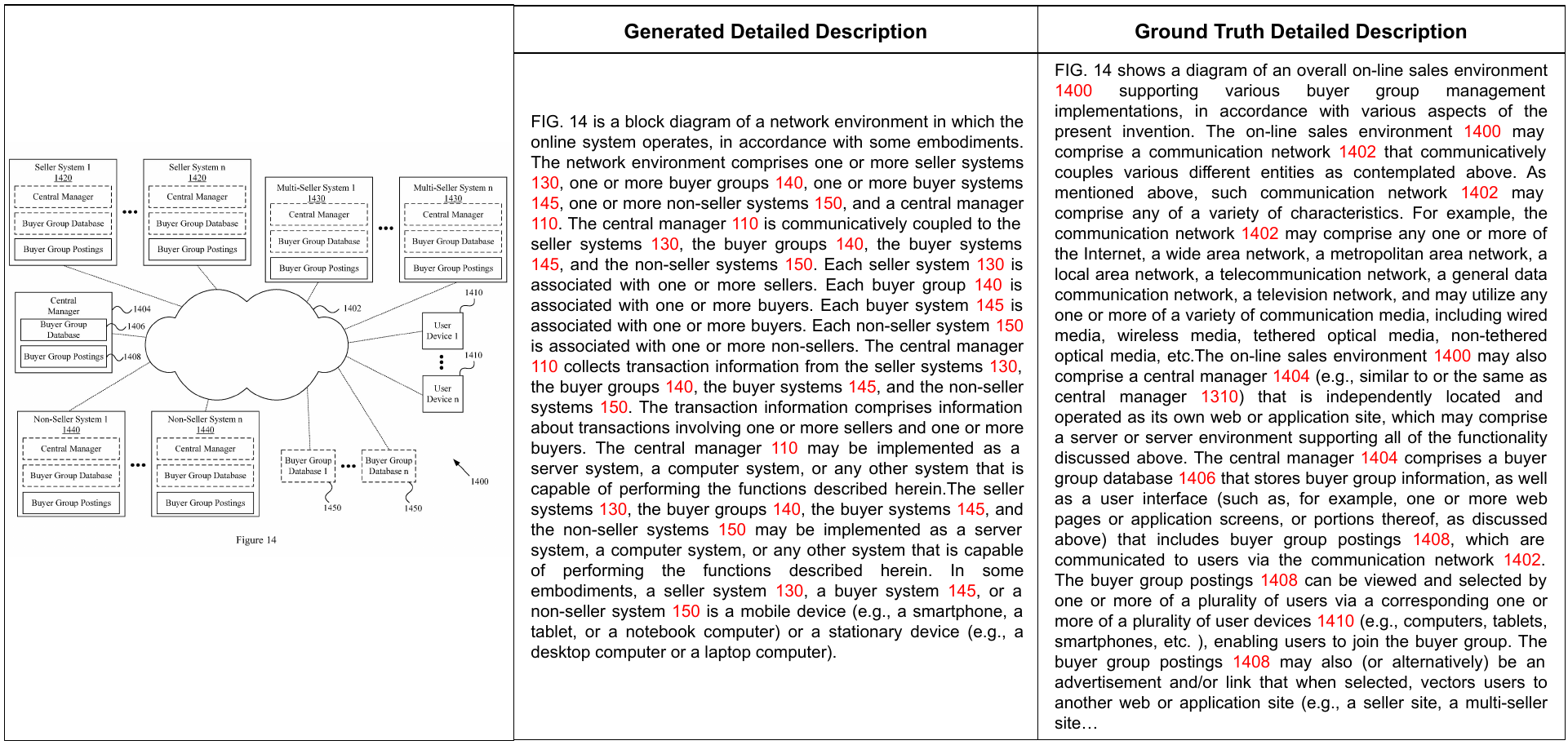}
\caption{Failure Case for the detailed description generated by \model{}. The text in red highlights the hallucinated node labels.}
\label{fig:detailed_failure_case}
\end{figure*}

These three error categories are also encountered in detailed descriptions generated by our model. Additionally, we identify two more failure cases in the detailed descriptions generated by our \model{} model, as shown in Fig.~\ref{fig:detailed_failure_case2}, which occurs when the node labels are associated incorrectly with concepts presented in nodes, and in Fig.~\ref{fig:detailed_failure_case}, which happens when the model hallucinates node labels. These cases usually occur due to the distortion of node labels or wiggly arrows connecting the nodes and node labels, during downsampling of images before being passed to PatentMME. 

\section{Future directions to mitigate failures}
In this section, we describe potential approaches to address the observed failure cases in our model’s generated descriptions. Specifically, we focus on reducing hallucinations and inaccuracies arising from missing contextual information, including figure references and technical details.

\noindent \textbf{Document-level Reasoning for better Cross-Figure References}
Our analysis reveals that the model occasionally hallucinates cross-figure references (as illustrated in our case studies (Appendix D.1)), particularly when an invention’s component is illustrated across multiple diagrams. To mitigate this issue, we propose enhancing document-level reasoning by linking interdependent figures throughout the patent document. By enabling the model to track and reconcile components and their relations across multiple figures, we can ensure that figure references are more accurate and context-aware. 

\noindent \textbf{Incorporation of External Knowledge Bases}
When key textual cues are absent or insufficient within a given figure, the model may hallucinate technical details. To address this limitation, we suggest integrating external technical knowledge sources—such as domain-specific knowledge bases or authoritative patent databases—into the generation process. By drawing on these external resources, the model can retrieve and incorporate accurate relevant information rather than hallucinating it. To this end, techniques inspired by Retrieval-Augmented Generation (RAG) can be employed to query large, domain-specific repositories and return the most relevant knowledge snippets. This retrieval step provides a verifiable grounding for the generated descriptions, significantly reducing the likelihood of hallucinated technical content.

\noindent \textbf{Grounded Description Generation}
Combining the above strategies can lead to a unified approach for grounded description generation, where figure content is consistently anchored to both internal textual references and external knowledge bases. By ensuring that each descriptive element corresponds to verifiable information, we can produce more reliable, less hallucinated, and overall higher-quality figure descriptions. This combination of document-level reasoning and external resource integration is a crucial step towards generating robust and trustworthy descriptions of patent figures.

\end{document}